\documentclass[lettersize,journal]{IEEEtran}
\usepackage{amsmath,amsfonts}
\usepackage{algorithmic}
\usepackage{array}
\usepackage{textcomp}
\usepackage{stfloats}
\usepackage{url}
\usepackage{verbatim}
\usepackage{graphicx}
\usepackage{multirow}
\usepackage{booktabs}
\usepackage{diagbox}
\usepackage[table]{xcolor}
\usepackage{subcaption} % 导言区添加
\def\BibTeX{{\rm B\kern-.05em{\sc i\kern-.025em b}\kern-.08em
    T\kern-.1667em\lower.7ex\hbox{E}\kern-.125emX}}
\usepackage{balance}
\begin{document}
\title{VADMamba++: Efficient Video Anomaly Detection via Hybrid Modeling in Grayscale Space}
% Efficient Video Anomaly Detection via Color-Aware with Hybrid Mamba-Transformer Network
\author{Jihao Lyu, Minghua Zhao, Jing Hu, Yifei Chen, Shuangli Du, Cheng Shi 
\thanks{The authors are with the Shaanxi Key Laboratory for Network Computing and Security Technology, School of Computer Science and Engineering, Xi'an University of Technology, Xi’an, 710048, China. Corresponding author: Minghua Zhao. zhaominghua@xaut.edu.cn}}

\markboth{Journal of \LaTeX\ Class Files,~Vol.~14, No.~8, August~2021}%
{Shell \MakeLowercase{\textit{et al.}}: A Sample Article Using IEEEtran.cls for IEEE Journals}

\maketitle

\begin{abstract}
VADMamba pioneered the introduction of Mamba to Video Anomaly Detection (VAD), achieving high accuracy and fast inference through hybrid proxy tasks. Nevertheless, its heavy reliance on optical flow as auxiliary input and inter-task fusion scoring constrains its applicability to a single proxy task. In this paper, we introduce VADMamba++, an efficient VAD method based on the Gray-to-RGB paradigm that enforces a Single-Channel to Three-Channel reconstruction mapping, designed for a single proxy task and operating without auxiliary inputs. This paradigm compels inferring color appearances from grayscale structures, allowing anomalies to be more effectively revealed through dual inconsistencies between structure and chromatic cues. Specifically, VADMamba++ reconstructs grayscale frames into the RGB space to simultaneously discriminate structural geometry and chromatic fidelity, thereby enhancing sensitivity to explicit visual anomalies. We further design a hybrid modeling backbone that integrates Mamba, CNN, and Transformer modules to capture diverse normal patterns while suppressing the appearance of anomalies. Furthermore, an intra-task fusion scoring strategy integrates explicit future-frame prediction errors with implicit quantized feature errors, further improving accuracy under a single task setting. Extensive experiments on three benchmark datasets demonstrate that VADMamba++ outperforms state-of-the-art methods while meeting performance and efficiency, especially under a strict single-task setting with only frame-level inputs.

\end{abstract}


\section{Introduction}
\label{sec:intro}

Video anomaly detection (VAD) aims to effectively identify abnormal events in videos while reducing manual inspection effort and false alarm rates. As it plays a crucial role in public safety, VAD requires both high detection accuracy and real-time performance \cite{liu2023generalized,lyu2025vadmamba}.

Since anomalous events are rare, unsupervised learning from abundant normal patterns is typically preferred, also framed as \textit{One-Class Classification} \cite{liu2023generalized,gong2019memorizing,sun2024dual}. The popular unsupervised VAD methods generally detect anomalies using two proxy tasks: frame- or object-level reconstruction \cite{singh2024attention,qiu2024video} and prediction \cite{wang2023video,yang2023video}. Advanced methods achieve these tasks through complex preprocessing, such as random masking \cite{ristea2024self}, Gaussian noise \cite{lyu2025moba, wu2023dss}, temporal inversion \cite{yang2025video}, optical flow \cite{cheng2023spatial,liu2021hybrid}, and difference guidance \cite{fang2020multi,zhong2025two}, to improve performance. Moreover, conventional methods adopt an RGB-to-RGB reconstruction paradigm~\cite{huang2024long,lyu2025bidirectional}, which allows models to learn appearance-level consistency, but often results in unnecessary color modeling for abnormal. In this work, inspired by image colorization~\cite{cong2024automatic}, which estimates RGB colors from grayscale images, we extend VAD to a Gray-to-RGB reasoning paradigm that infers appearance from grayscale structure, thereby preventing redundant color reproduction of anomalies.

\begin{figure}[t]
	\centering
	%\includegraphics[width=0.8\linewidth]{egfigure.eps}
	\includegraphics[width=\linewidth]{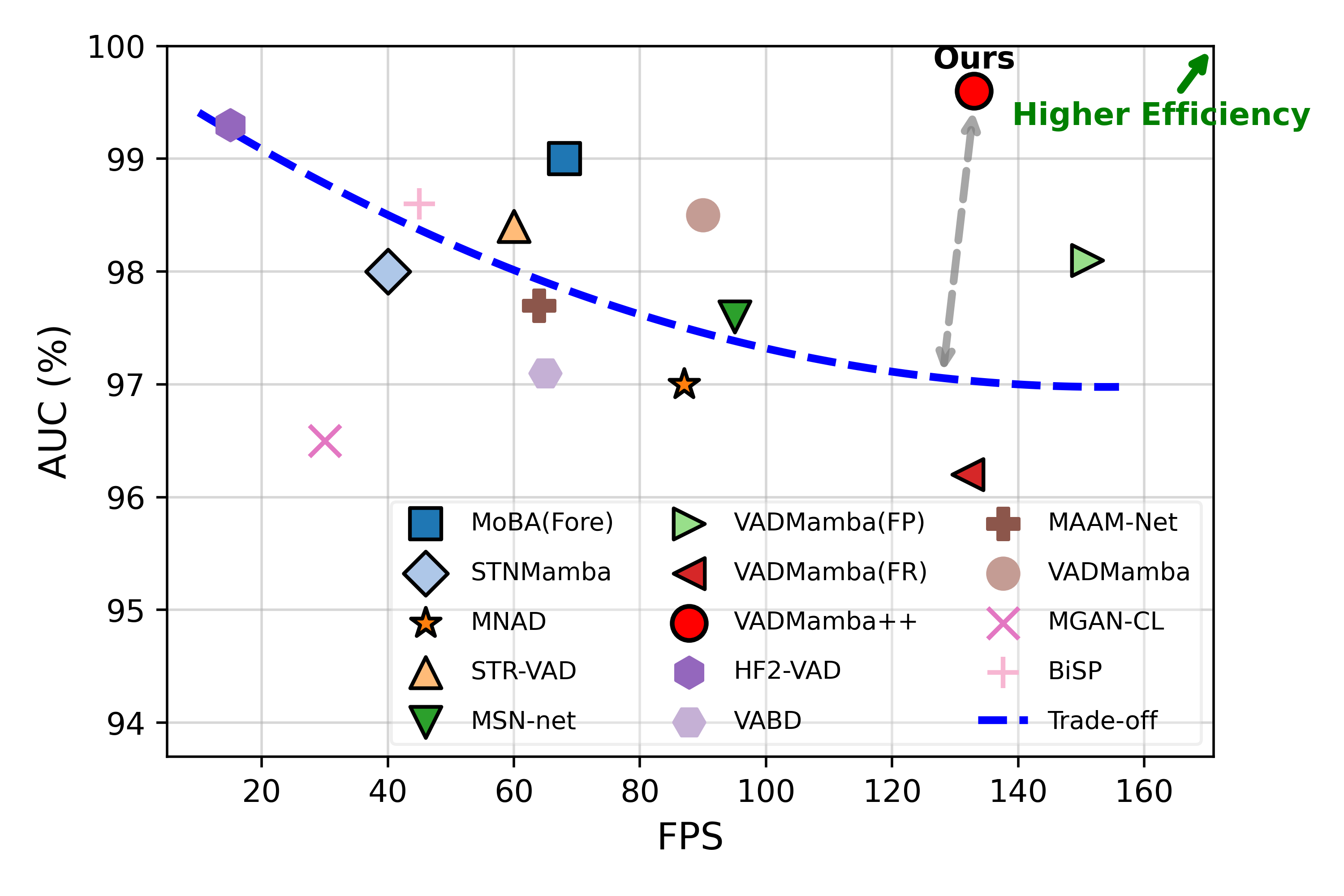}
	\caption{
		%		Comparison of FPS and AUC with state-of-the-art methods on Ped2. The blue dashed line denotes the performance trade-off curve, revealing the general trend that higher accuracy typically leads to lower speed. Our proposed VADMamba++ notably breaks this trade-off.
		Comparison of AUC and FPS on Ped2~\cite{sabokrou2015real}, where single-task methods are marked with black edges. The \textbf{\textcolor{blue}{blue dashed line}} shows the typical tradeoff: higher accuracy often reduces speed, while our VADMamba++ notably breaks this trend.
		% The blue dashed line indicates the Pareto frontier,
	}
	\label{fig:f1}
\end{figure}

Additionally, in most existing methods focus on improving accuracy while neglecting inference efficiency. As shown in Figure \ref{fig:f1}, the blue dashed line illustrates the conventional trend in VAD, where higher accuracy is typically achieved at the cost of lower speed. For instance, HF2-VAD~\cite{liu2021hybrid}, which relies on optical flow and object-level inputs, attains higher accuracy but suffers from slower inference. Although VADMamba~\cite{lyu2025vadmamba} demonstrates the practical potential of Mamba for VAD by balancing detection accuracy and inference speed, it still struggles to manage the tradeoff between hybrid and single proxy tasks. Despite advancements of VADMamba \cite{lyu2025vadmamba}, its performance improvements are primarily attributed to three aspects: hybrid proxy tasks requiring complex coordination, auxiliary inputs introducing optical-flow overhead, and inter-task fusion scoring that delays anomaly feedback. Consequently, these gaps call for a more efficient single-task setting to achieve a better balance accuracy and speed.

To fill these gaps, we propose VADMamba++, which achieves a desirable trade-off between detection accuracy and inference speed through a single proxy task. Figure \ref{fig:f2} compares the pipelines of VADMamba (RGB-to-RGB) and VADMamba++ (Gray-to-RGB). Unlike VADMamba, which emphasizes appearance reconstruction, VADMamba++ reframes anomaly detection as a reasoning task: it takes grayscale single-channel images to predict RGB three-channel outputs, compelling the model to learn scene structures and regenerate plausible appearances, where reasoning failures naturally emerge as strong indicators of anomalies. Specifically, VADMamba++ introduces a novel colorization-based reasoning paradigm that force our method to infer chromatic representations from structural cues and thereby amplify anomaly scores arising, a novel hybrid modeling backbone integrating Mamba-based bidirectional SS2D for efficient sequence reasoning, CNN-based Dual Pre-activation for local spatial refinement, and Transformer blocks for global context aggregation, and a novel intra-task anomaly scoring strategy that simplifies score computation while maintaining detection accuracy.

Our main contributions can be summarized as follows:

\begin{figure}[t]
	\centering
	\includegraphics[width=\linewidth]{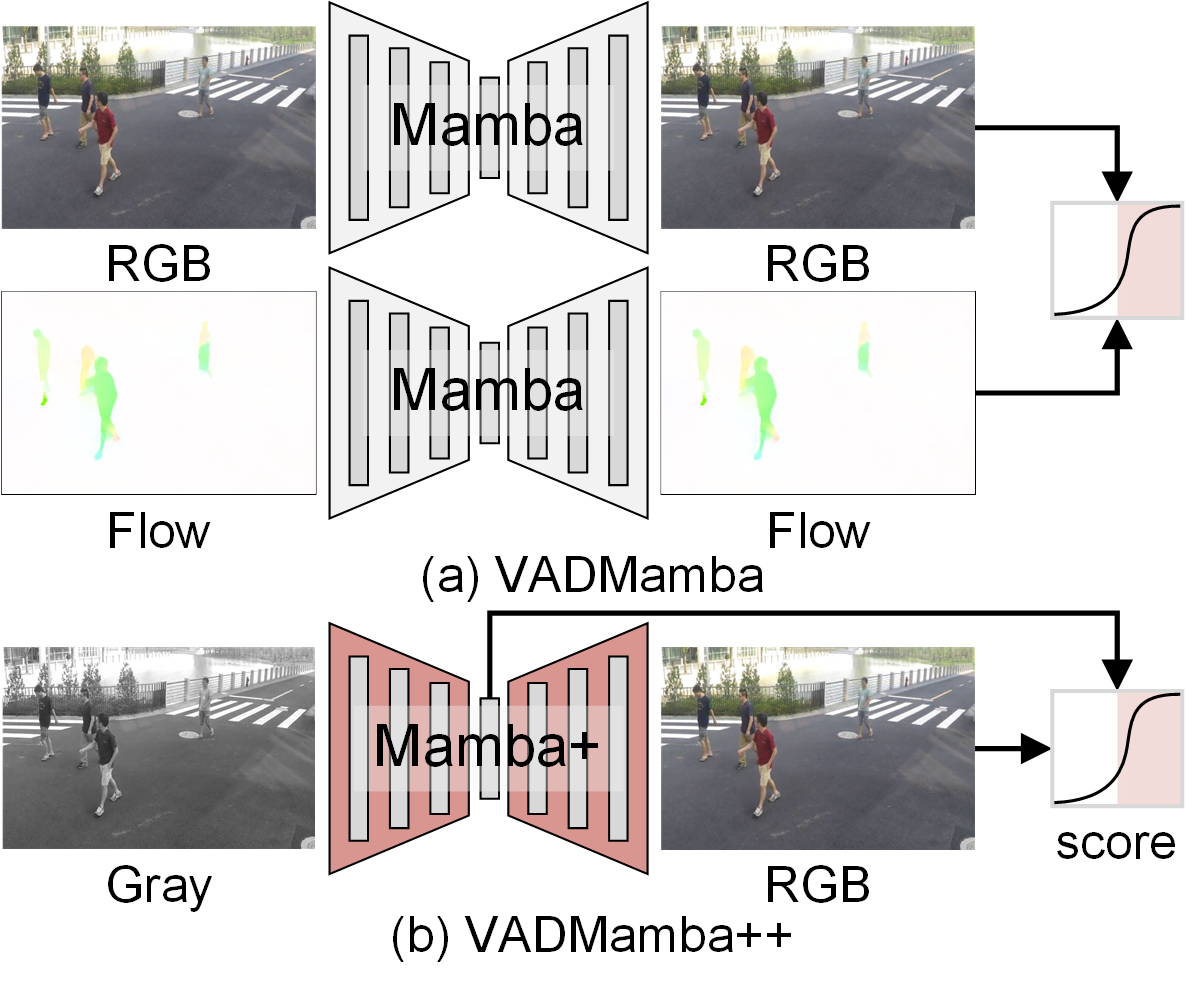}
	\caption{Pipeline comparison between (a) VADMamba and (b) VADMamba++. We highlight three key evolutions: 1) Multi-input \emph{vs.} \textbf{Single-input}; 2) Multi-task \emph{vs.} \textbf{Single-task}; and 3) Inter-task fusion \emph{vs.} \textbf{Intra-task fusion}.}
	\label{fig:f2}
\end{figure}
\begin{itemize}
	\item We propose VADMamba++, a frame prediction method based on a Gray-to-RGB reasoning paradigm to balance accuracy and speed via a single proxy task.
	\item We introduce a hybrid Transformer–Mamba–CNN network that facilitates exploratory integration across the encoding and decoding stages.
	\item We perform intra-task fusion anomaly scoring during inference by adaptively weighting explicitly reconstructed errors and implicitly quantified feature errors.
	\item Our VADMamba++ comprehensively improves upon VADMamba, achieving competitive state-of-the-art performance across three benchmark datasets, particularly excelling in single proxy tasks with frame-only inputs.
\end{itemize}
\noindent{\textbf{Difference from the original conference version (i.e., VADMamba \cite{lyu2025vadmamba}).}} Compared to the original VADMamba, our VADMamba++ in this work transitions from an RGB-to-RGB reconstruction to a Gray-to-RGB reasoning paradigm, which compels the model to infer chromatic cues from grayscale structures without relying on auxiliary optical flow inputs. We replace the multi-task design with a single-task framework powered by a hybrid Transformer-Mamba-CNN backbone that integrates global context, long-range temporal dynamics, and local spatial refinement. Furthermore, we introduce an intra-task fusion scoring strategy that adaptively weights explicit prediction errors and implicit quantized feature errors. These architectural optimizations significantly reduce the number of parameters and FLOPs while achieving superior detection accuracy and a robust inference speed.

\section{Related Work}
Unlike supervised VAD datasets such as UCF-Crime~\cite{sultani2018real} and XD-Violence~\cite{wu2020not}, which contain video- and frame-level labels, unsupervised VAD employs unlabeled datasets. The core challenge of unsupervised VAD lies in modeling normal patterns and quantifying deviations as anomaly scores \cite{liu2023generalized}. Using unlabeled datasets such as Ped2~\cite{sabokrou2015real}, Avenue~\cite{lu2013abnormal}, and ShanghaiTech~\cite{luo2017revisit}, most unsupervised VAD methods capture latent regularities through prediction, reconstruction, or feature embedding, thereby enabling robust anomaly detection \cite{kim2025mpe}.

\textbf{Detection Modeling Architecture} in unsupervised VAD is critical for capturing spatio-temporal features and guiding how they are extracted, fused, and utilized for anomaly detection. In recent years, a variety of architectures have been explored, including CNN-based, Transformer-based, and Mamba-based models, each providing distinct mechanisms for feature representation and anomaly detection. \textbf{\textit{1) CNN-based.}} CNNs are widely applied in the vision community, effectively extracting image features and capturing spatial patterns. Specifically, most VAD methods utilize the AutoEncoder (AE) \cite{gong2019memorizing,park2020learning,lyu2025moba,yang2022dynamic}, GAN \cite{singh2024attention,li2023multi,sun2024dual,liang2024c,wu2023dss} and Diffusion \cite{zhou2025video,liu2024vadiffusion,yan2023feature} to detection anomaly. Building on this, prior work further pursued high-precision performance through bidirectional prediction \cite{lyu2025bidirectional,zhong2022bidirectional}, hybrid methods \cite{hu2025noise,liu2021hybrid,huang2024long,zhong2025two}, and self-supervised methods \cite{wu2023dss, park2025fast}. \textbf{\textit{2) Transformer-based.}} Unlike CNNs with limited local receptive fields, the landmark Transformer \cite{vaswani2017attention} models global dependencies, enabling it to capture long-term temporal and cross-spatial relationships in videos or images. Some methods \cite{qiu2024video, le2025hstforu} for consecutive videos typically employ ViTs to extract features during the encoding stage, while integrating CNNs for single-frame decoding. Additionally, other works \cite{yang2023video, huang2022hierarchical} fully leverage ViTs to extract temporal features, achieving efficiency through optimized input processing and model implementation. Moreover, since ViTs operate on image patches, some methods \cite{ristea2024self, madan2023self} leverage masking to reduce computational cost while improving accuracy. \textbf{\textit{3) Mamba-based.}} Recently, Mamba \cite{gu2023mamba} has offered new directions for vision tasks, achieving scalable long-sequence modeling and effective temporal dependency capture, unlike CNNs and Transformers. VADMamba \cite{lyu2025vadmamba} was the first VAD to incorporate Mamba, achieving a favorable balance between performance and efficiency through a multi-proxy task design, followed by STNMamba \cite{li2024stnmamba}, which integrated RGB difference features and multiple memory models for frame prediction. Beyond the weakly supervised VAD \cite{xiao2025multilingual}, other studies \cite{liu2024vmamba,hemambaad} pioneered the application of Mamba. 

Despite these architectural advances, how to achieve complementary advantages across different frameworks and explore optimal training strategies for hybrid architectures remains an open challenge.

%hu2022detecting tran2024transformer

\begin{figure*}[ht]
	\centering
	\includegraphics[width=\linewidth]{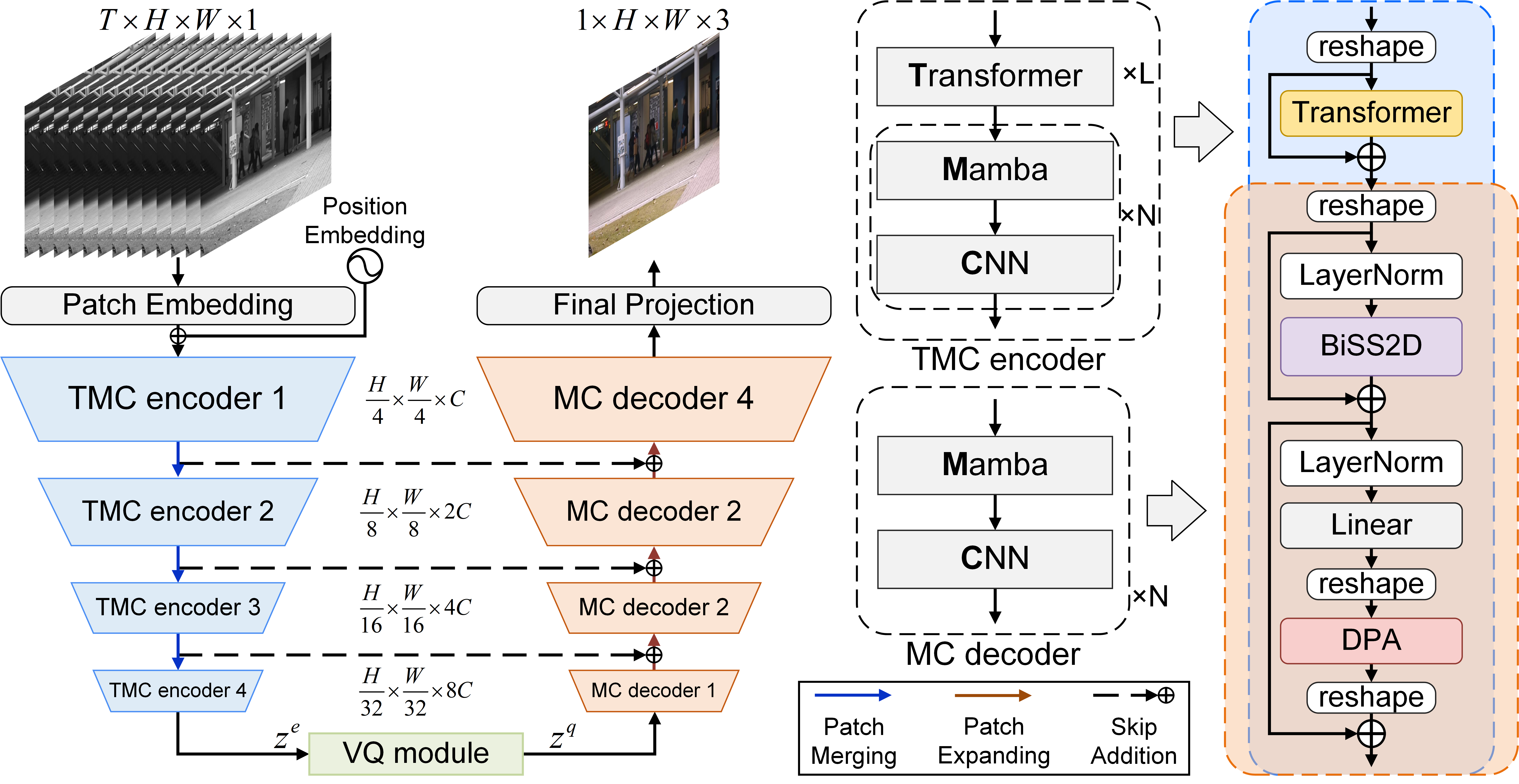}
	\caption{Overview of the proposed VADMamba++, which employs an asymmetric Mamba-based encoder–decoder framework to colorize grayscale input frames into the next RGB frame. The TMC encoder combines Transformer, Mamba, and CNN blocks for spatiotemporal feature extraction, whereas the MC decoder reconstructs a colorized frame guided by the quantized representation from the VQ module.}
	\label{fig:f3}
\end{figure*}

\textbf{Anomaly Score Representation} provides a unified metric for quantifying deviations from normal patterns. It can be further categorized into explicit reconstruction-based and implicit distance-based methods, which capture  deviations from different representational perspectives. \textbf{\textit{1) Explicit Reconstruction-based.}} Current reconstruction-, prediction-, and hybrid-based VAD methods follow a unified paradigm: generating a complete reconstructed or predicted frame and quantifying their deviation from the ground truth using traditional metrics such as L1/L2 distance \cite{gong2019memorizing,yan2023feature} or PSNR \cite{hu2025noise,kommanduri2024dast,zhang2022hybrid}. Most single-task VAD methods evaluate anomalies using traditional scores, while advanced methods further refine anomaly scoring based on error maps through multi-scale pyramids \cite{zhong2022bidirectional,zhong2023associative,lyu2025moba}, multi-frame prediction error \cite{kim2025mpe} and maximum local error \cite{cao2024context}, achieving more accurate anomaly scoring. Recently, SSIM-based evaluation \cite{fan2025ssim} has been shown to outperform traditional metrics, while LLM-based semantic scoring \cite{gao2025suvad} assess anomalies by comparing textual descriptions of normal and abnormal events. \textbf{\textit{2) Implicit Distance-based.}} Unlike explicit reconstruction, implicit discrimination methods detect anomalies directly from high-dimensional features without incurring substantial additional inference time. Distance-based evaluation on implicit representations is the basic method \cite{ahn2024videopatchcore}, which has subsequently evolved to the Log-likelihood \cite{hirschorn2023normalizing,wu2025flow}, feature hashing \cite{zhang2016video,lu2022learnable}, and Gaussian mixture model-based \cite{ micorek2024mulde,wang2025icc}. Moreover, CRCL \cite{liu2025crcl} models normal video patterns via causal mechanisms and derives anomaly scores from clustering distances among causal factors. Building upon this, some methods \cite{park2020learning, zhang2024multi,cao2024context,huang2024long,liu2023msn} further integrate explicit and implicit paradigms for more comprehensive VAD. 
%, cho2022unsupervised han2024mutuality zhuvision  huang2023video

However, these score metrics share a common limitation: they primarily focus on appearance-based cues while neglecting deeper color reasoning, particularly the consideration of feature discrepancies within the latent feature space.

\section{Preliminary}

\textbf{Problem Definition.} Given an unlabeled training video sequence $S=\{I_t\}_{t=1}^{N}$ consisting exclusively of normal events, let $I_{1:T} \in \mathbb{R}^{T \times H \times W \times 3}$ denote a video clip of $T$ consecutive RGB frames. We define a grayscale transformation $\mathcal{G}: \mathbb{R}^{3} \to \mathbb{R}^{1}$ that maps the RGB sequence to its single-channel structural counterpart $x_{1:T} = \mathcal{G}(I_{1:T})$, where $x_{1:T} \in \mathbb{R}^{T \times H \times W \times 1}$ represents the input grayscale images. The objective is to learn the mapping from the grayscale structural domain to the color domain to predict the target color frame $I_{T+1} \in \mathbb{R}^{1 \times H \times W \times 3}$.

\textbf{State Space Models.} Our VADMamba++ is built upon the Mamba \cite{gu2023mamba}, specifically the Selective Scan (S6) mechanism. Conceptually, S6 models the temporal evolution of a system where a 1-D input stream $x(t) \in \mathbb{R}$ drives a high-dimensional latent state $h(t) \in \mathbb{R}^N$. This latent state effectively compresses historical context to project the final output $y(t) \in \mathbb{R}$. Formally, the continuous-time evolution of this system is defined by the differential equation:
\begin{equation}
\begin{aligned}
h'(t) &= \mathbf{A}h(t) + \mathbf{B}x(t), \\
y(t) &= \mathbf{C}h(t),
\end{aligned}
\label{eq:ssm_continuous}
\end{equation}
where $\mathbf{A} \in \mathbb{R}^{N \times N}$, $\mathbf{B} \in \mathbb{R}^{N}$, and $\mathbf{C} \in \mathbb{R}^{N}$ represent learnable projection parameters.

To accommodate the input data discretization characteristics of deep learning models while maintaining the differentiability of linear ordinary differential equations, S6 introduces a sampling step $\Delta$ and uses the zero-order hold principle to discretize the matrices $\mathbf{A}$ and $\mathbf{B}$. Consequently, Eq.~\ref{eq:ssm_continuous} is re-written to:
\begin{equation}
\begin{aligned}
h_k &= \overline{\mathbf{A}} h_{k-1} + \overline{\mathbf{B}} x_k, \\
y_k &= \overline{\mathbf{C}} h_k,
\end{aligned}
\label{eq:ssm_discrete}
\end{equation}
where $\overline{\mathbf{A}} = \exp(\Delta \mathbf{A})$, $\overline{\mathbf{B}} = (\Delta \mathbf{A})^{-1}(\exp(\Delta \mathbf{A}) - \mathbf{I}) \cdot \Delta \mathbf{B}$, and $\overline{\mathbf{C}}$ denotes the discretized learnable matrix. Additionally, S6 dynamically parameterizes matrices $\mathbf{A}$ and $\mathbf{B}$ based on inputs to effectively emphasize critical information while suppressing irrelevant details.

%Additionally, S6 dynamically parameterizes the matrices $\mathbf{A}$ and $\mathbf{B}$, allowing the model to adapt these parameters to the input data. This mechanism enables S6 to effectively emphasize important information while suppressing irrelevant details, which is crucial for capturing anomalies in video sequences.

%To accommodate the input data discretization characteristics of deep learning models while maintaining the differentiability of linear ordinary differential equations, S6 introduces a sampling step $\Delta$ and uses the zero-order hold principle to discretize the matrices $\mathbf{A}$ and $\mathbf{B}$. Consequently, Eq.~\ref{eq:ssm_continuous} is re-written to:
%\begin{equation}
%\begin{aligned}
%h_k &= \overline{\mathbf{A}} h_{k-1} + \overline{\mathbf{B}} x_k, \\
%y_k &= \overline{\mathbf{C}} h_k,
%\end{aligned}
%\label{eq:ssm_discrete}
%\end{equation}
%where $\overline{\mathbf{A}} = e^{\Delta \mathbf{A}}$, $\overline{\mathbf{B}} = (e^{\Delta \mathbf{A}} - \mathbf{I})\mathbf{A}^{-1}\mathbf{B}$, and $\overline{\mathbf{C}}$ denotes the discretized learnable matrix (typically $\overline{\mathbf{C}} \approx \mathbf{C}$).
%
%In addition, S6 dynamically parameterizes the parameter matrices $\mathbf{A}$ and $\mathbf{B}$ to allow the model to adjust these parameters according to different input data adaptively. This way, S6 can effectively emphasize important information while ignoring irrelevant details, which is crucial for capturing anomalies in video sequences.

\section{Methodology}
Figure \ref{fig:f3} illustrates the overall architecture of the proposed VADMamba++, a U-Net framework enhanced with Mamba blocks. By applying grayscale preprocessing to the input video sequence, the model simultaneously performs future-frame colorization and prediction. During inference, the anomaly score is computed jointly computed from explicit prediction frame errors and the implicit quantized feature errors to quantify the abnormality degree of each frame.

\subsection{VADMamba++}
To improve the model’s sensitivity to anomalous events and emphasize anomaly-related errors, we propose a Gray-to-RGB (G2R) modeling VAD framework that integrates the efficient feature extraction of Mamba blocks, the local spatial representation ability of CNNs, and the strong temporal modeling capability of Transformers, thereby achieving a balance between detection accuracy and inference efficiency. As illustrated in Figure \ref{fig:f3}, VADMamba++ comprises three main components: the TMC encoder, the MC decoder, and a vector quantization (VQ) bottleneck as in VADMamba \cite{lyu2025vadmamba}. 

The TMC encoder first extracts features from the grayscale inputs with positional embeddings via a patch embedding operation. The encoding sequence consists of Transformer blocks, Mamba-based bidirectional SS2D (BiSS2D) modules, and CNN-based Dual Pre-activation (DPA) modules. Except for the last layer, Patch Merge down-sampling is performed after each coding stage. The resulting high-dimensional features $z_e$ are subsequently processed by the VQ bottleneck to perform feature compression and codebook retrieval, producing the quantized representation $z_q$. The MC decoder differs from the encoder primarily by removing the Transformer module. Since the decoder only needs to generate a single frame, it emphasizes reconstructing spatial details rather than modeling temporal dependencies. Except for the first layer, Patch Expansion up-sampling is performed before each decoding stage. After feature extraction through a four-stage encoder–decoder pipeline, the final projection layer generates the predicted color frame. The encoder–decoder design of VADMamba++, built upon a G2R reasoning paradigm, enables effective aggregation of multi-frame temporal information within the encoder while minimizing redundant computation in the decoder, yielding a more compact and efficient overall architecture.

\subsection{Encoder-Decoder with Mamba+}
In the following, we mainly present three key modules of the encoder and decoder: Transformer blocks, Mamba-based BiSS2D, and CNN-based DPA.

\textbf{Transformer.} Within the encoder, each Transformer block adopts the standard multi-head self-attention ($\text{MHSA}$) and feed-forward network ($\text{FFN}$) structure. Given an input feature sequence $x_{v} \in \mathbb{R}^{T \times N \times d}$, the output $y$ is computed as:
\begin{equation}
\begin{aligned}
x' &= x_{v} + \text{MHSA}(\text{LN}(x_{v})), \\
y  &= x' + \text{FFN}(\text{LN}(x')),
\end{aligned}
\end{equation}
where $\text{LN}$ denotes layer normalization.

\begin{figure}[t]
	\centering
	\includegraphics[width=\linewidth]{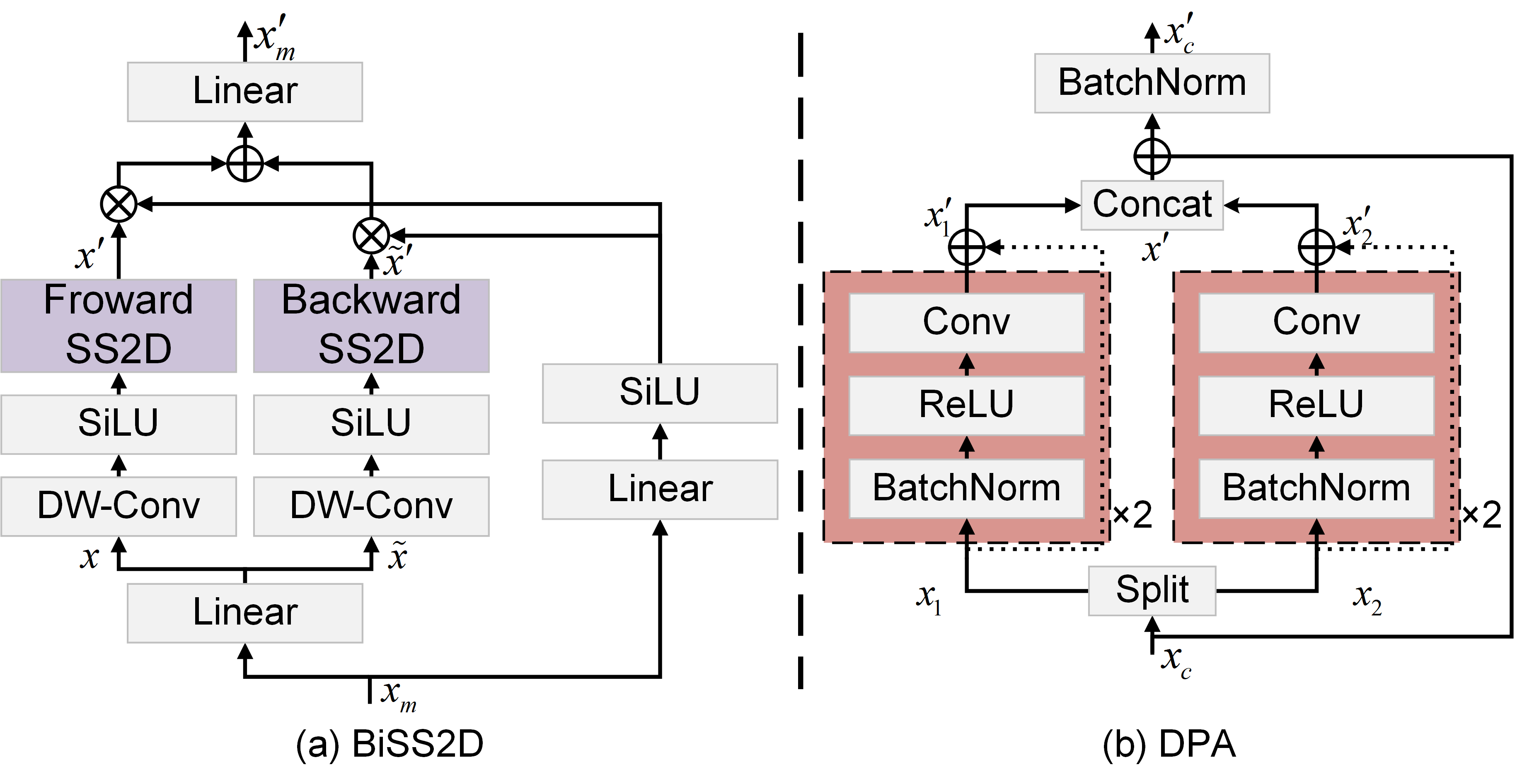}
	\caption{The architectures of (a)BiSS2D and (b)DPA.}
	\label{fig:f4}
\end{figure}

%\noindent{}
\textbf{Bidirectional SS2D.} To enhance the structural–chromatic consistency and improve the contextual integrity of the RGB space, based on SS2D \cite{liu2024vmamba}, we propose a BiSS2D module that establishes symmetric spatial dynamics through dual forward–backward state transitions. 
As shown in Figure~\ref{fig:f4}(a), given an input feature $x_m \in \mathbb{R}^{H \times W \times C}$, we obtain two complementary embeddings:
\begin{equation}
x, \bar{x} = \text{Linear}(x_m), \quad \tilde{x} = \text{Flip}(x),
\end{equation}
where $\tilde{x}$ is the spatially reversed version of $x$, enabling reversed-state propagation. A gating variable $z$ is derived from $\bar{x}$ to regulate the subsequent fusion process.

\begin{equation}
z = \text{SiLU}(\text{Linear}(\bar{x})).
\end{equation}

Each branch applies depthwise convolution and nonlinearity before the directional SS2D operators:
\begin{equation}
\begin{aligned}
x' &= \text{SS2D}_{\text{fwd}}\!\big(\text{SiLU}(\text{DWConv}(x))\big), \\
\tilde{x}' &= \text{SS2D}_{\text{bwd}}\!\big(\text{SiLU}(\text{DWConv}(\tilde{x}))\big).
\end{aligned}
\end{equation}

The bidirectional states are fused via gated modulation and feature coupling:
\begin{equation}
x_m' = \text{Linear}\!\big((x' \odot z) + (\tilde{x}' \odot z )\big),
\end{equation}

%\noindent{}
\textbf{Dual Pre-activation.} As shown in Figure~\ref{fig:f4}(b), we design a DPA module to improve feature extraction efficiency through a lightweight dual-path structure. By dividing channels into two complementary subspaces, DPA enables parallel feature transformation and reduces redundant computation. DPA first splits the input $x_c \in \mathbb{R}^{C \times H \times W}$ into two channel-wise parts:

\begin{equation}
x_1, x_2 = \text{Split}(x_c),
\end{equation}
where $x_1, x_2 \in \mathbb{R}^{\frac{C}{2} \times H \times W}$. 
Each branch applies two stacked pre-activation modules following the order of Batch Normalization (BN), ReLU, and Convolution (Conv):
\begin{equation}
\mathcal{F}_i(x) = \text{Conv}_1\!\big(\sigma(\text{BN}(\text{Conv}_3(\sigma(\text{BN}(x_i)))))\big),
\end{equation}
where $\text{Conv}_3$ and $\text{Conv}_1$ denote convolutional layers with kernel sizes of $3\times3$ and $1\times1$, respectively. Then, a residual coupling is preserved within each subspace to maintain identity flow:
\begin{equation}
x_i' = \mathcal{F}_i(x_i) + x_i, \quad i \in \{1,2\}.
\end{equation}

The resulting subspace outputs are concatenated and re-normalized before being merged with the global identity mapping:
\begin{equation}
x_c' = \text{BN}\!\big(\text{Concat}(x_1', x_2') + x_c\big).
\end{equation}

\subsection{Training and Inference}

\textbf{Training loss.} To efficiently train VADMamba++ for high-quality color frame prediction, we jointly optimize multiple loss functions, including prediction loss $\mathcal{L}_{pred}$, gradient loss $\mathcal{L}_{grad}$, color-awere loss $\mathcal{L}_{color}$, and VQ loss $\mathcal{L}_{vq}$.
\begin{equation}\label{eq1}
\mathcal{L}_{pred}=\left\|I_{T+1}-\hat{I}_{T+1}\right\|_{2},
\end{equation} 
where $I_{T+1}$ and $\hat{I}_{T+1}$ denote the ground truth and the predicted frame.

\begin{equation}\label{eq:2}
\begin{gathered}
\mathcal{L}_{grad}=\sum_{i,j} \left \| \left | I_{T+1}^{i,j}-I_{T+1}^{i-1,j} \right | - \left | \hat{I}_{T+1}^{i,j}-\hat{I}_{T+1}^{i-1,j} \right | \right \|_{1} + \\ \left \| \left | I_{T+1}^{i,j-1}-I_{T+1}^{i,j} \right | -\left | \hat{I}_{T+1}^{i,j-1}-\hat{I}_{T+1}^{i,j} \right | \right \|_{1},
\end{gathered}
\end{equation}
where $i$ and $j$ denotes the spatial index of the frame.
\begin{equation}
\mathcal{L}_{color} = \sum_{i=1}^{2} \left\| \psi_i(I_{T+1}) - \psi_i(\hat{I}_{T+1}) \right\|_1
\end{equation}
where $\psi_i(\cdot)$ denotes the $i$-th ReLU layer feature of the VGG \cite{simonyan2014very}. Detailed VQ loss $\mathcal{L}_{vq}$ in \cite{lyu2025vadmamba}. The total training loss is as follows:
\begin{equation}
\mathcal{L}=\mathcal{L}_{pred}+\mathcal{L}_{grad}+\mathcal{L}_{color}+\mathcal{L}_{vq}. 
\end{equation}

\textbf{Inference score.} The anomaly score for each frame is computed by combining the explicit reconstruction error $S_E$ and the implicit quantized feature error $S_I$. The explicit error $S_E$ is computed between the ground truth frame $I_t$ and the predicted frame $\hat{I}_t$:
\begin{equation} 
S_E = \frac{1}{N} \sum_{1}^{N} \|I_{t} - \hat{I}_{t}\|_2,
\end{equation}
where $N$ denotes the total number of pixels. Likewise, the implicit error $S_I$ between quantized feature $z_q$ and original feature $z_e$ is computed as the top-$k$ averaged $L_2$ distance:
\begin{equation}\label{eq:k}
S_I = \frac{1}{k} \sum_{1}^{k} \text{Topk}_i \!\left( \| z_q - z_e \|_2 \right),
\end{equation}
where $\text{Topk}_i(\cdot)$ indicates the $i$-th largest value. Before fusion, we first compute the PSNR-based scores for both errors, followed by normalization and Gaussian smoothing to maintain temporal consistency. Finally, the overall anomaly score $S$ is obtained through weighting:
\begin{equation}\label{eq:l}
S = \text{PSNR}(S_E) + \lambda \cdot \text{PSNR}(S_I),
\end{equation}
where $\text{PSNR}(\cdot)$ denotes the Peak Signal-to-Noise Ratio score and $\lambda$ is a tradeoff parameter of $S_I$. 

%Finally, the PSNR is normalized to the range $[0,1]$ using Eq.~(\ref{eq6}) and further smoothed with a 1D Gaussian filter to generate the final anomaly score.
%
%\begin{equation}\label{eq6}
%S(I_{t})=\frac{\operatorname{PSNR}(I_{t}, \hat{I}_{t})-\min (\operatorname{PSNR}(I_{t}, \hat{I}_{t}))}{\max (\operatorname{PSNR}(I_{t}, \hat{I}_{t}))-\min(\operatorname{PSNR}(I_{t}, \hat{I}_{t}))}.
%\end{equation}

% Prior to fusion, both scores are normalized and then smoothed with a Gaussian filter to maintain temporal consistency. Finally, the overall anomaly score $S$ is obtained as a weighted combination of the two PSNR-based scores:

\section{Experiments}
\textbf{Experimental Setup.} We evaluate VADMamba++ on Ped2~\cite{sabokrou2015real}, Avenue~\cite{lu2013abnormal}, and ShanghaiTech (SHT)~\cite{luo2017revisit}. Following previous works~\cite{liu2021hybrid,park2020learning,lyu2025vadmamba,yan2023feature}, we adopt the Area Under the ROC Curve (AUC) as the evaluation metric. The input frames are normalized to $[-1,1]$ and resized into $224 \times 224$. Models are trained with AdamW ($\text{lr}=2e-4$) using an input length of $T=16$, with $L=1$ and $N=1$ for all datasets. All experiments are implemented in PyTorch on a NVIDIA 4090 GPU. For the grayscale Ped2, we enforce the Channel from 1C to 3C transformation, compelling the model to reconstruct pseudo-RGB targets from single-channel inputs.

%For Ped2 dataset, the inputs and outputs are single-channel and three-channel images, respectively. 

\begin{table}[t]
	\centering
	\caption{Comparison with state-of-the-art methods on three benchmark datasets. We list the feature of the methods, in which 'F' denotes frame, 'OF' denotes optical flow, 'FG' denotes foreground, 'FD' denotes frame differential, 'SA' denotes synthetic anomaly. $^{\ast}$ denotes the same GPU device. The \colorbox{gray!20}{gray} marker denotes Mamba-based methods. }
	\label{tab:t1}
	
	\begin{subtable}[t]{\linewidth}
		\centering		
		\caption{Comparison with single-task methods. The best results in each category are marked in \textbf{bold}. $^{\dagger}$ denotes trained on patch-level input.}
		\setlength{\tabcolsep}{1.7mm}{
		\begin{tabular}{lccccc}
			\toprule
			Method & Feature & Avenue & SHT & Ped2 & FPS\\   
			\midrule
			MESDnet \cite{fang2020multi} &  F+FD   & 86.3 & 73.2 & 95.6 &  - \\
			STGCN-FFP \cite{cheng2023spatial}  &  F+OF   & 88.4 & 73.7 & 96.9 &  - \\
			MoBA (Flow) \cite{lyu2025moba} &  F+OF   & 88.7 & 75.2 & 98.4 &  68$^{\ast}$ \\
			MoBA (Fore) \cite{lyu2025moba} &  F+FG   & 89.6 & 74.0 & \textbf{99.0} &  68$^{\ast}$ \\
			MA-PDM (OF) \cite{zhou2025video} &   F$^{\dagger}$+OF$^{\dagger}$   & 88.7 & - & 93.7 &  25 \\
			MA-PDM (FD) \cite{zhou2025video} &   F$^{\dagger}$+FD$^{\dagger}$   & \textbf{91.3} & \textbf{79.2} & 98.6 &  45 \\
			\cellcolor{gray!20}STNMamba \cite{li2024stnmamba}  &  \cellcolor{gray!20}F+FD   & \cellcolor{gray!20}89.0 & \cellcolor{gray!20}74.9 & \cellcolor{gray!20}98.0 & \cellcolor{gray!20}40  \\
			\midrule
			MemAE \cite{gong2019memorizing}   & F       & 83.3 & 71.2 & 94.1 & 31$^{\ast}$\\
			MNAD  \cite{park2020learning}   & F      & 88.5 & 70.5 & 97.0 & 87$^{\ast}$\\
			MSN-net \cite{liu2023msn}  & F      & 89.4 & 73.4 & 97.6 & 95  \\
			DSS-Net  \cite{wu2023dss} & F      & 90.6 & 75.5 & 97.2 & -\\
			STR-VAD \cite{wang2023video} &  F & 86.1 & 73.2 & 98.4 & 60  \\
			USTN-DSC  \cite{yang2023video}  &  F   & 89.9 & 73.8 & 98.1 & - \\ 
			DAST-Net \cite{kommanduri2024dast}& F  & 89.8 & 73.7 & 97.9 & 22  \\
			A2D-GAN \cite{singh2024attention}& F  & 91.0 & 74.2 & 97.4 & -  \\
			FastAno+ \cite{park2025fast} &  F   & 87.8 & 75.2 & 98.1 & 130 \\ 
			STEAL \cite{fan2025ssim} &  F   & 87.5 & 76.4 & 98.9 & 75	 \\ 
			\cellcolor{gray!20}VADMamba (FP)  \cite{lyu2025vadmamba}  &  \cellcolor{gray!20}F  & \cellcolor{gray!20}86.2 & \cellcolor{gray!20}74.4 & \cellcolor{gray!20}98.1 & \cellcolor{gray!20}151$^{\ast}$\\ 	
			\cellcolor{gray!20}VADMamba (FR)  \cite{lyu2025vadmamba}  &  \cellcolor{gray!20}OF  & \cellcolor{gray!20}87.3 & \cellcolor{gray!20}72.2 & \cellcolor{gray!20}96.2 & \cellcolor{gray!20}132$^{\ast}$\\ 
			\cellcolor{gray!20}VADMamba++ & \cellcolor{gray!20}F & \cellcolor{gray!20}\textbf{91.9} & \cellcolor{gray!20}\textbf{77.1} & \cellcolor{gray!20}\textbf{99.6} & \cellcolor{gray!20}133$^{\ast}$ \\ 
			\bottomrule
		\end{tabular}
		}
		\label{tab:1a}
	\end{subtable} 
	\hfill
	\vspace{0.3 mm}
	\begin{subtable}[t]{\linewidth}
		\centering
		\caption{Comparison with multi-task methods. We list the number of tasks for the methods. $^{\ddagger}$ denotes trained on object-level input. '( )' denotes the improvement over VADMamba++.}
		\setlength{\tabcolsep}{1.7mm}{
			\begin{tabular}{lcccccc}
				\toprule
				Method & Feature & Task & Avenue & SHT & Ped2 & FPS\\   
				\midrule
				HF2-VAD  \cite{liu2021hybrid}& F$^{\ddagger}$+OF$^{\ddagger}$  &  2   & 91.1 & 76.2 & 99.3 & 15$^{\ast}$\\
				VABD \cite{li2021variational} & F+OF  &  2   & 86.6 & 78.2(1.1) & 97.1 &  65$^{\ast}$\\
				AMMC-Net \cite{cai2021appearance}  & F+OF  &  2   & 86.6 & 73.7 & 96.6 & - \\
				MAAM-Net \cite{wang2023memory}   & F+OF & 2    & 90.9 & 71.3 & 97.7 & 64 \\
				BiFI \cite{deng2023bi}  & F+OF   &  2   & 89.7 & 75.0 & 98.9 &  - \\
				AED-MAE \cite{ristea2024self}  & F+SA   &  2   & 91.3 & 79.1(2.0) & 95.4 &  - \\
				\cellcolor{gray!20}VADMamba \cite{lyu2025vadmamba}  &  \cellcolor{gray!20}F+OF & \cellcolor{gray!20}2 & \cellcolor{gray!20}91.5 & \cellcolor{gray!20}77.0 & \cellcolor{gray!20}98.5 & \cellcolor{gray!20}90$^{\ast}$\\ 
				\midrule
				SSAE \cite{cao2024scene}& F$^{\ddagger}$  &  2   & 90.2 & 80.5(3.4) & - & 10$^{\ast}$\\
				SLM \cite{shi2023video}& F$^{\ddagger}$  &  3   & 90.9 & 78.8(1.7) & 97.6 & - \\
				LLSH \cite{lu2022learnable}   &  F  &  2    & 88.6 & 77.6(0.5) & 91.3 & 25$^{\ast}$\\
				SSAGAN \cite{huang2022self}  & F  &  2   & 88.8 & 74.3 & 97.6 & 40 \\
				MGAN-CL \cite{li2023multi}  & F   &  2   & 87.1 & 73.6 & 96.5 &  30 \\
				FPDM \cite{yan2023feature} & F &   2    & 90.1 & 78.6(1.5) & - & 8  \\
				VADiffusion \cite{liu2024vadiffusion} &  F &  2   & 87.2 & 71.7 & 98.2 & 2 \\ 
				GroupGAN  \cite{sun2024dual}  & F   &  2   & 85.5 & 73.1 & 96.6 &  70 \\
				C$^{2}$Net  \cite{liang2024c} &  F  &   2 & 87.5 & 71.4 & 98.0 & -  \\ 
				PDM-Net \cite{huang2024long}& F   &  2 & 88.1 & 74.2 & 97.7 & -  \\ 
				BiSP \cite{lyu2025bidirectional}& F   &  2 & 89.5 & 76.4 & 98.6 & 45$^{\ast}$ \\ 
				TSMDC \cite{zhong2025two}& F   &  3 & 90.6 & 74.4 & 98.2 & -  \\ 
				\cellcolor{gray!20}VADMamba++ & \cellcolor{gray!20}F & \cellcolor{gray!20}\textbf{1} & \cellcolor{gray!20}\textbf{91.9} & \cellcolor{gray!20}\textbf{77.1} & \cellcolor{gray!20}\textbf{99.6}& \cellcolor{gray!20}\textbf{133$^{\ast}$} \\ 
				\bottomrule
			\end{tabular} 
		}
		\label{tab:1b}
	\end{subtable}
\end{table}

\textbf{Quantitative results.} We compare VADMamba++ with state-of-the-art unsupervised VAD methods, including both single-task and multi-task settings, as summarized in Table~\ref{tab:t1}, which also reports method input features and inference FPS. Table~\ref{tab:1a} shows that: (1) VADMamba++ achieves the highest performance across all three benchmarks, attaining 91.9\% on Avenue, 77.1\% on SHT, and 99.6\% on Ped2 under the single-task setting, while exhibiting 133 FPS. Considering both AUC and FPS, it demonstrates the best overall tradeoff. (2) Compared with hybrid input methods, VADMamba++ delivers superior accuracy alongside a remarkable inference speed. (3) Relative to the single-task variant of VADMamba, VADMamba++ maintains high FPS without compromising accuracy. Table~\ref{tab:1b} indicates multi-task methods that: (1) Some methods~\cite{li2021variational, cao2024scene} achieve slightly higher scores on the SHT dataset but suffer from lower inference speeds, typically below 100~FPS, which limits their practical applicability. In contrast, VADMamba++ attains competitive accuracy at 133~FPS. (2) Compared with other multi-task methods, VADMamba++ achieves comparable or superior accuracy while maintaining the highest FPS, thereby demonstrating a favorable balance between efficiency and performance. (3) These results highlight the strong advantages of VADMamba++ in both efficiency and accuracy, making it highly suitable for real-time anomaly detection. Moreover, with only 14.3M parameters and 1.9G FLOPs, VADMamba++ offers a compact and efficient design, outperforming VADMamba (29.5M, 5.2G) and its variant FP (14.8M, 2.6G). Overall, VADMamba++ achieves an excellent tradeoff, validating the feasibility and effectiveness of the proposed reasoning paradigm, while delivering superior detection performance and strong potential for real-time deployment.

\begin{figure*}[t]
	\centering
	\includegraphics[width=\linewidth]{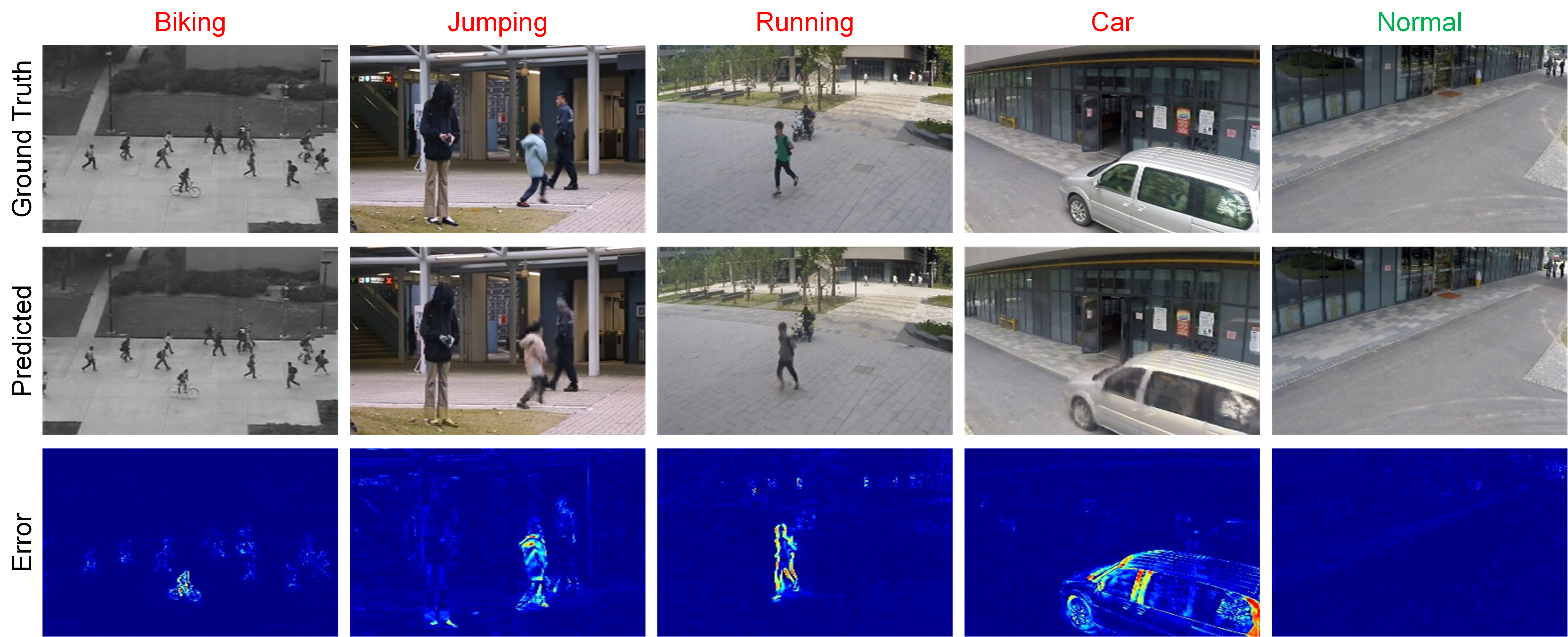}
	\caption{Examples of ground truth frames, predicted results, and error maps across five scenarios. The first four columns depict abnormal cases, whereas the last column presents a normal case. Brighter regions in the error maps indicate larger prediction errors.}
	\label{fig:f7}
\end{figure*}

\textbf{Qualitative results.} In Figure~\ref{fig:f7}, we present the predicted frames and error maps of VADMamba++ across four anomaly scenarios, with a normal scenario shown in the fifth column. In all four anomaly types, the abnormal regions consistently exhibit higher reconstruction errors, even though the inputs in Column 1 are grayscale. The colored anomalies presented in Columns 2 to 4 cannot be accurately recovered in the RGB space, which can be attributed to the G2R reasoning paradigm. In contrast, the normal background colors remain unaffected, and the normal case in Column 5 further validates this conclusion.

\begin{figure}[ht]
	\centering
	\includegraphics[width=\linewidth]{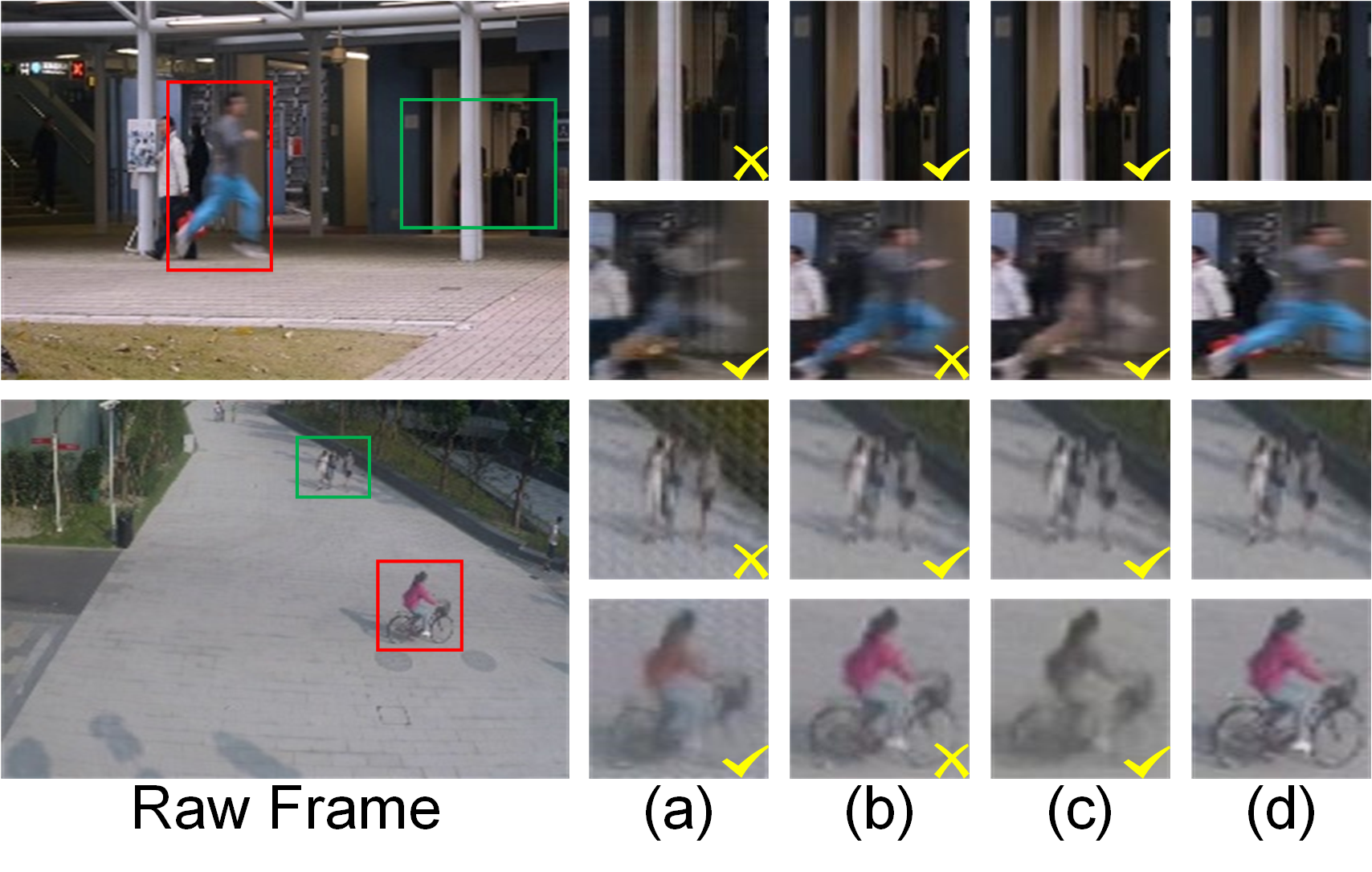}
	\caption{Qualitative comparison of normal (green box) and anomaly (red box) results across different methods. (a) VADMamba, (b) VADMamba++ (w/o G2R), (c) VADMamba++, and (d) Ground Truth object.}	\label{fig:f7.1}
\end{figure}

As shown in Figure~\ref{fig:f7.1}, the compared methods exhibit distinct reconstruction behaviors for normal (green box) and abnormal (red box) regions. VADMamba produces noticeable errors for abnormal but fails to preserve the structural cues of normal, and VADMamba++ (w/o Gray) enhances texture fidelity, yet its RGB-to-RGB reconstruction lacks the G2R dual explicit discrimination. In contrast, VADMamba++ faithfully reconstruction of color and structure in normal regions while amplifying chromatic and structural deviations in abnormal ones, yielding larger reconstruction errors that better highlight anomalies.

\begin{figure}[h]
	\centering
	\includegraphics[width=\linewidth]{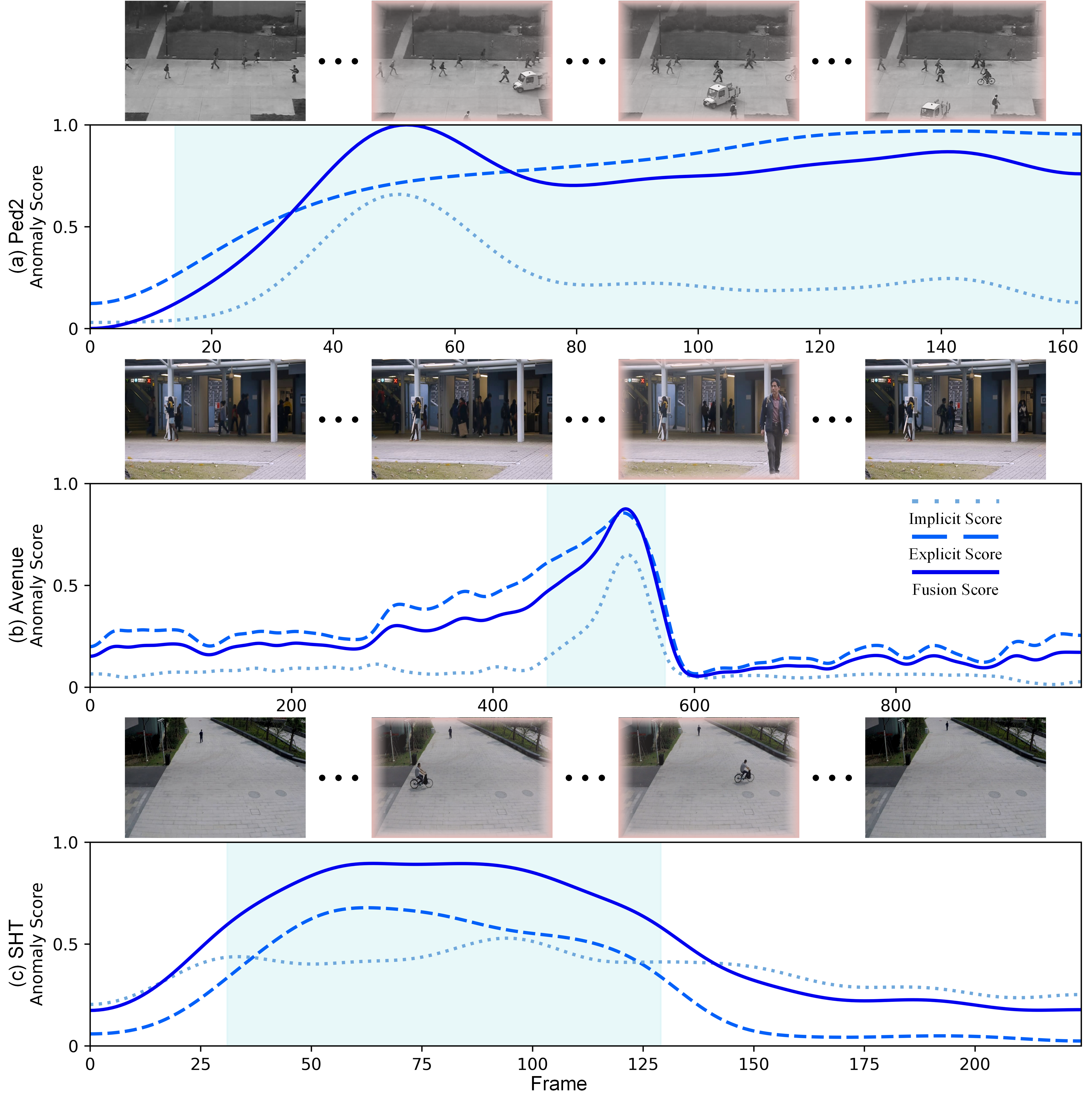}
	\caption{Frame-level anomaly scores on (a) Ped2, (b) Avenue, and (c) SHT. Red boxes denote abnormal frames, while dotted, dashed, and solid lines indicate implicit, explicit, and fusion scores, respectively.}
	\label{fig:f5}
\end{figure}

Figure~\ref{fig:f5} shows the anomaly scores, the results across three experimental groups exhibit consistent discriminative patterns under different scenarios, with the SHT dataset demonstrating the most pronounced performance. The fusion score presents the clearest and highest response peaks during anomalous events and maintains smooth and continuous variations within abnormal intervals, outperforming the fluctuations observed in the explicit and implicit branches. This indicates that the fusion mechanism can more accurately capture the temporal continuity of abnormal events, showing stronger sequential modeling capability in complex scenarios such as SHT, where anomalies are prolonged and background interference is significant.

\begin{figure}[t]
	\centering
	\includegraphics[width=\linewidth]{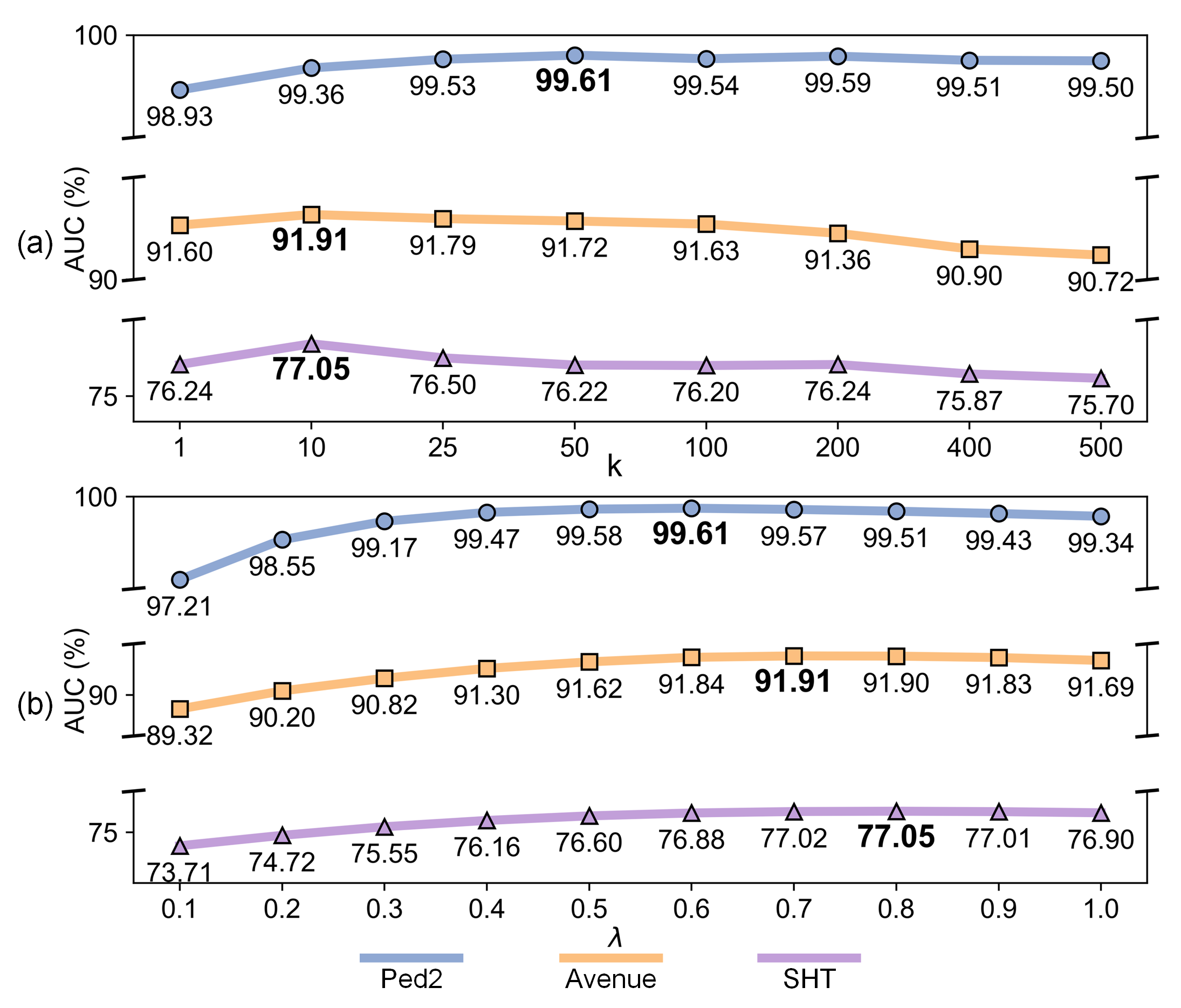}
	\caption{Sensitivity analysis of (a) $k$ and (b) $\lambda$ on three datasets.}
	\label{fig:f6}
\end{figure}

\textbf{Parameter sensitivity.} To evaluate the robustness of the hybrid score, we analyze the sensitivity of $k$ in Eq.~\ref{eq:k} and $\lambda$ in Eq.~\ref{eq:l}, as shown in Fig.~\ref{fig:f6}. The parameter $k$ represents the number of top-ranked implicit features considered during inference. On Ped2, larger $k$ values lead to slightly higher AUC owing to stable image quality and minimal illumination variation. In contrast, Avenue and SHT achieve optimal performance at $k=10$, where a smaller $k$ better handles illumination fluctuations and multi-scene diversity. The computational difference between $k=10$ and $k=500$ is negligible, thus exerting no influence on inference efficiency. The coefficient $\lambda$ controls the contribution of the implicit score, where moderate values strike an effective balance between explicit and implicit errors, and the overall variation remains small. Overall, the consistent trends across $k$ and $\lambda$ indicate that the proposed framework is robust and insensitive to hyperparameter variations.

\textbf{Ablation study.} To validate the effectiveness of VADMamba++, we conduct comprehensive ablations studies, and analyze the contributions of key components, architectural design, module ordering, and loss functions. Notably, for fairness, $k=200$ and the adaptive $\lambda$ is applied to all ablation results. 

As shown in Table~\ref{tab:t3}, the results demonstrate that each component is indispensable to our model’s effectiveness. The chromatic reasoning task notably enhances color-sensitive representation learning and improves anomaly exposure, whereas RGB reconstruction (w/o G2R) fails to be applicable. Notably, the grayscale Ped2 lacks chromatic cues, and redundant three-channel mapping impedes learning. The Transformer (T), BiSS2D, and DPA modules provide complementary strengths in capturing global and local spatio-temporal dependencies, while the $S_{I}$ integration strategy effectively fuses explicit and implicit errors for more robust scoring.

\begin{table}[h]
	\centering
	\caption{Ablation studies on the component.}
	\label{tab:t3}
	
	\begin{tabular}{cccccc}
		\toprule
		w/o & G2R & T & BiSS2D & DPA & $S_{I}$ \\
		\midrule
		Ped2 & 98.5 & 98.4 & 96.2 & 98.2 & 98.1  \\
		Avenue & 89.8 & 89.8 & 89.0 & 89.4 & 88.9   \\
		SHT & 74.3 & 74.7 & 73.6 & 74.2 &  73.5  \\
		\bottomrule
	\end{tabular}
\end{table}

As shown in Table~\ref{tab:t4}, in Ped2, the asymmetric configuration with the Transformer in the encoder (E) only achieves the best performance, confirming that global context modeling benefits feature encoding, while a lightweight decoder (D) better handles single-frame prediction. Varying the number of $N$ and $L$ further shows that a compact setting ($N{=}1$, $L{=}1$) yields optimal results, as deeper structures add complexity without improving accuracy.

\begin{table}[h]
	\centering
	\caption{Impact of encoder-decoder feature extraction on Ped2.}
	\label{tab:t4}
	
	\begin{subtable}[t]{0.47\linewidth}
		\centering		
		\caption{Impact of Transformer (T).}
		\vspace{0.55em}
		\begin{tabular}{ccc}
			\toprule
			\diagbox[width=4em]{E}{D} & w/o T & w/ T \\
			\midrule
			w/o T & 98.4 & 98.6 \\
			w/ T  & \textbf{99.6} & 98.7 \\
			\bottomrule
		\end{tabular}
		
		\label{tab:t4a}
	\end{subtable}
	\hfill
	\begin{subtable}[t]{0.51\linewidth}
		\centering
		\caption{Impact of different $N$ and $L$.}
		\setlength{\tabcolsep}{1.7mm}{
			\begin{tabular}{cccc}
				\toprule
				\diagbox[width=4em]{$N$}{$L$}& 1 & 2 & 3 \\
				\midrule
				1 & \textbf{99.6} & 98.5 & 98.2   \\
				2 & 98.4 & 98.8 & 98.6   \\
				3 & 97.9 & 98.4 & 97.8   \\
				\bottomrule
			\end{tabular}
		}
		\label{tab:t4b}
	\end{subtable}
\end{table}

As shown in Table~\ref{tab:t5}, the \textbf{T}ransformer$\rightarrow$\textbf{M}amba$\rightarrow$\textbf{C}NN (TMC) arrangement achieves the highest accuracy across all datasets, validating a hierarchical flow where the Transformer captures global context, Mamba models long-range temporal dynamics, and CNN refines local spatial details. Alternative orderings disrupt this coarse-to-fine progression and lead to reduced performance.

\begin{table}[h]
	\centering
	\caption{Impact of varying module ordering in the encoder.}
	\label{tab:t5}
	
	\begin{tabular}{ccccc}
		\toprule
		& TMC & TCM & MCT& CMT\\
		\midrule
		Ped2 & \textbf{99.6}  & 98.9  & 98.8  &  98.6  \\
		Avenue & \textbf{91.4}  &  90.7 & 90.5  &  89.8  \\
		SHT &  \textbf{76.2} & 75.4  &  74.9 &  74.3  \\
		\bottomrule
	\end{tabular}
	
\end{table}

Table~\ref{tab:t6} shows that progressively adding $L_{\text{vq}}$, $L_{\text{grad}}$, and $L_{\text{color}}$ to the baseline $L_{\text{pred}}$ steadily improves performance, with total gains of +1.2/+2.5/+3.1 on three datasets. These enhancements reflect complementary supervision: $L_{\text{vq}}$ refines latent representations, $L_{\text{grad}}$ preserves structural details, and $L_{\text{color}}$ enforces chroma consistency, together forming a balanced and effective training objective.

\begin{table}[h]
	\centering
	\caption{Effectiveness of loss functions.}
	\label{tab:t6}
	
	\begin{tabular}{ccccccc}
		\toprule
		$\mathcal{L}_{pred}$ & $\mathcal{L}_{vq}$ & $\mathcal{L}_{grad}$ & $\mathcal{L}_{color}$ & Ped2 & Avenue & SHT  \\
		\midrule
		\checkmark &  &  &  & 98.2 & 88.9 &  73.1 \\
		\checkmark & \checkmark &  &  & 98.4 & 89.2 & 73.6   \\
		\checkmark & \checkmark & \checkmark &  & 98.6 & 89.8 &  74.4  \\
		\checkmark & \checkmark & \checkmark & \checkmark & \textbf{99.6} & \textbf{91.4} & \textbf{76.2}   \\
		\bottomrule
	\end{tabular}
\end{table}

The ablation studies collectively validate the design of VADMamba++ with the G2R paradigm, hybrid TMC architecture with optimal asymmetry and ordering, compact depth, and multi-term supervision jointly contribute to high accuracy and fast speed.

As shown in Table~\ref{tab:t7}, increasing the input frame length $T$ from 4 to 8 consistently boosts performance, indicating that moderate temporal expansion enhances motion perception and stabilizes spatiotemporal reasoning. The AUC slightly fluctuates at $T=12$ but reaches its optimum at $T=16$ with 99.4\%, demonstrating that this length provides the most effective balance between temporal richness and feature compactness. When $T$ increases to 20, the AUC drops to 98.9\%, suggesting that overly long sequences introduce redundant temporal cues and degrade representation efficiency. This trend converges at $T=16$, aligning with the observations in VADMamba, which confirms a similar optimal input length for effective temporal modeling.

\begin{table}[h]
	\centering
	\caption{The AUC for different values of $T$ on the Ped2 dataset.}
	\label{tab:t7}
	\begin{tabular}{lccccc}
		\toprule
		$T$   & 4 & 8 & 12   & 16  & 20 \\
		\midrule
		AUC(\%) $\uparrow$ & 98.7  & 99.1  & 99.3 & \textbf{99.6}  & 99.1  \\
		\bottomrule
	\end{tabular}
\end{table}

As shown in Table~\ref{tab:t8}, the integration of the G2R paradigm and the Implicit score from VADMamba++ consistently enhances the baseline VADMamba (FP) across all benchmarks. The G2R paradigm alone yields modest but stable improvements, particularly on Avenue (+0.6) and SHT (+0.3), indicating that grayscale-guided reasoning helps refine appearance modeling. The Implicit score brings more significant gains, especially on Avenue (+1.1), suggesting that implicit feature alignment effectively captures deeper semantic consistency. When both modules are jointly applied, the performance reaches the highest AUC of 98.9\%, 88.0\%, and 75.3\% on Ped2, Avenue, and SHT, respectively, demonstrating a clear complementary effect between explicit grayscale reasoning and implicit representation learning. This further proves the effectiveness of the proposed VADMamba++.

\begin{table}[h]
	\centering
	\caption{Incremental evaluation of VADMamba with the G2R paradigm and the Implicit score.}
	\label{tab:t8}

		\begin{tabular}{lccc}
			\toprule
			& Ped2 & Avenue  & SHT \\
			\midrule
			VADMamba (FP) \cite{lyu2025vadmamba} &    98.1  &  86.2   &  74.4   \\
			+ G2R   &   98.2 (+0.1)  &   86.8 (+0.6)  &   74.7 (+0.3) \\
			+ Implicit     &  98.8 (+0.6)    &  87.3 (+1.1)   & 74.9 (+0.5)    \\
			+ G2R \& Implicit &  98.9 (+0.7)    &  88.0 (+1.8)   & 75.3 (+0.9)    \\
			
			\bottomrule
		\end{tabular}
	
\end{table}

%More detailed experiments and analysis can be found in the \textbf{\textit{supplementary materials}}.

\section{Conclusion}
In this paper, we design a Gray-to-RGB reasoning paradigm to propose an efficient VAD, which learns to reconstruct the normal color space while failing to recover the abnormal. Our proposed VADMamba++ employs hybrid modeling, leveraging explicit and implicit errors to achieve higher efficiency anomaly detection. Therefore, operating under a single proxy task and without any auxiliary inputs, it achieves a better tradeoff between accuracy and speed. Extensive experiments on three benchmark datasets demonstrate the superiority of our method over existing single- and multi-task methods.


\section*{Acknowledgments}
This work was supported by the Natural Science Foundation of Shaanxi Province, China (2024JC-ZDXM-35), National Natural Science Foundation of China (No.62571425), and Doctoral Dissertation Innovation Fund of Xi’an University of Technology (BC202621).

\bibliographystyle{IEEEtran}
\bibliography{references.bib}

@article{sun2024dual,
	title={Dual GroupGAN: An unsupervised four-competitor (2V2) approach for video anomaly detection},
	author={Sun, Zhe and Wang, Panpan and Zheng, Wang and Zhang, Meng},
	journal={Pattern Recognition},
	volume={153},
	pages={110500},
	year={2024},
	publisher={Elsevier}
}

@article{liu2024vadiffusion,
	title={Vadiffusion: Compressed domain information guided conditional diffusion for video anomaly detection},
	author={Liu, Hao and He, Lijun and Zhang, Miao and Li, Fan},
	journal={IEEE Transactions on Circuits and Systems for Video Technology},
	volume={34},
	number={9},
	pages={8398--8411},
	year={2024},
	publisher={IEEE}
}

@article{li2021variational,
	title={Variational abnormal behavior detection with motion consistency},
	author={Li, Jing and Huang, Qingwang and Du, Yingjun and Zhen, Xiantong and Chen, Shengyong and Shao, Ling},
	journal={IEEE Transactions on Image Processing},
	volume={31},
	pages={275--286},
	year={2021},
	publisher={IEEE}
}

@article{xiao2025multilingual,
	title={Multilingual-Prompt-Guided Directional Feature Learning for Weakly Supervised Video Anomaly Detection},
	author={Xiao, Chizhuo and Xiao, Yang and Zhou, Joey Tianyi and Fang, Zhiwen},
	journal={IEEE Transactions on Pattern Analysis and Machine Intelligence},
	year={2025},
	publisher={IEEE}
}

@article{kim2025mpe,
	title={MPE: Multi-frame prediction error-based video anomaly detection framework for robust anomaly inference},
	author={Kim, Yujun and Kim, Young-Gab},
	journal={Pattern Recognition},
	volume={164},
	pages={111595},
	year={2025},
	publisher={Elsevier}
}

@inproceedings{huang2024long,
	title={Long short-term dynamic prototype alignment learning for video anomaly detection},
	author={Huang, Chao and Wen, Jie and Liu, Chengliang and Liu, Yabo},
	booktitle={Proceedings of the Thirty-Third International Joint Conference on Artificial Intelligence},
	pages={866--874},
	year={2024}
}

@inproceedings{cai2021appearance,
  title={Appearance-motion memory consistency network for video anomaly detection},
  author={Cai, Ruichu and Zhang, Hao and Liu, Wen and Gao, Shenghua and Hao, Zhifeng},
  booktitle={Proceedings of the AAAI conference on artificial intelligence},
  volume={35},
  number={2},
  pages={938--946},
  year={2021}
}

@article{zhang2022hybrid,
  title={Hybrid Attention and Motion Constraint for Anomaly Detection in Crowded Scenes},
  author={Zhang, Xinfeng and Fang, Jinpeng and Yang, Baoqing and Chen, Shuhan and Li, Bin},
  journal={IEEE Transactions on Circuits and Systems for Video Technology},
  year={2022},
  publisher={IEEE}
}

@inproceedings{zhou2025video,
	title={Video anomaly detection with motion and appearance guided patch diffusion model},
	author={Zhou, Hang and Cai, Jiale and Ye, Yuteng and Feng, Yonghui and Gao, Chenxing and Yu, Junqing and Song, Zikai and Yang, Wei},
	booktitle={Proceedings of the AAAI Conference on Artificial Intelligence},
	volume={39},
	number={10},
	pages={10761--10769},
	year={2025}
}

@article{wang2023memory,
  title={Memory-augmented appearance-motion network for video anomaly detection},
  author={Wang, Le and Tian, Junwen and Zhou, Sanping and Shi, Haoyue and Hua, Gang},
  journal={Pattern Recognition},
  volume={138},
  pages={109335},
  year={2023},
  publisher={Elsevier}
}

@inproceedings{liu2021hybrid,
  title={A hybrid video anomaly detection framework via memory-augmented flow reconstruction and flow-guided frame prediction},
  author={Liu, Zhian and Nie, Yongwei and Long, Chengjiang and Zhang, Qing and Li, Guiqing},
  booktitle={Proceedings of the IEEE/CVF international conference on computer vision},
  pages={13588--13597},
  year={2021}
}

@inproceedings{gong2019memorizing,
  title={Memorizing normality to detect anomaly: Memory-augmented deep autoencoder for unsupervised anomaly detection},
  author={Gong, Dong and Liu, Lingqiao and Le, Vuong and Saha, Budhaditya and Mansour, Moussa Reda and Venkatesh, Svetha and Hengel, Anton van den},
  booktitle={Proceedings of the IEEE/CVF International Conference on Computer Vision},
  pages={1705--1714},
  year={2019}
}

@inproceedings{park2020learning,
  title={Learning memory-guided normality for anomaly detection},
  author={Park, Hyunjong and Noh, Jongyoun and Ham, Bumsub},
  booktitle={Proceedings of the IEEE/CVF conference on computer vision and pattern recognition},
  pages={14372--14381},
  year={2020}
}

@article{cao2024scene,
	title={Scene-dependent prediction in latent space for video anomaly detection and anticipation},
	author={Cao, Congqi and Zhang, Hanwen and Lu, Yue and Wang, Peng and Zhang, Yanning},
	journal={IEEE transactions on pattern analysis and machine intelligence},
	year={2024},
	publisher={IEEE}
}

@article{huang2022self,
	title={Self-supervised attentive generative adversarial networks for video anomaly detection},
	author={Huang, Chao and Wen, Jie and Xu, Yong and Jiang, Qiuping and Yang, Jian and Wang, Yaowei and Zhang, David},
	journal={IEEE transactions on neural networks and learning systems},
	volume={34},
	number={11},
	pages={9389--9403},
	year={2022},
	publisher={IEEE}
}

@article{fang2020multi,
	title={Multi-encoder towards effective anomaly detection in videos},
	author={Fang, Zhiwen and Zhou, Joey Tianyi and Xiao, Yang and Li, Yanan and Yang, Feng},
	journal={IEEE Transactions on Multimedia},
	volume={23},
	pages={4106--4116},
	year={2020},
	publisher={IEEE}
}

@article{yang2025video,
	title={Video anomaly detection via self-supervised and spatio-temporal proxy tasks learning},
	author={Yang, Qingyang and Wang, Chuanxu and Liu, Peng and Jiang, Zitai and Li, Jiajiong},
	journal={Pattern Recognition},
	volume={158},
	pages={111021},
	year={2025},
	publisher={Elsevier}
}

@article{zhang2016video,
	title={Video anomaly detection based on locality sensitive hashing filters},
	author={Zhang, Ying and Lu, Huchuan and Zhang, Lihe and Ruan, Xiang and Sakai, Shun},
	journal={Pattern Recognition},
	volume={59},
	pages={302--311},
	year={2016},
	publisher={Elsevier}
}

@article{lyu2025moba,
	title={MoBA: Motion Memory-Augmented Deblurring AutoEncoder for Video Anomaly Detection},
	author={Lyu, Jiahao and Zhao, Minghua and Hu, Jing and Huang, Xuewen and Du, Shuangli and Shi, Cheng and Lv, Zhiyong},
	journal={Knowledge-Based Systems},
	pages={115218},
	year={2025},
	publisher={Elsevier}
}

@inproceedings{deng2023bi,
	title={Bi-directional frame interpolation for unsupervised video anomaly detection},
	author={Deng, Hanqiu and Zhang, Zhaoxiang and Zou, Shihao and Li, Xingyu},
	booktitle={Proceedings of the IEEE/CVF winter conference on applications of computer vision},
	pages={2634--2643},
	year={2023}
}

@article{zhong2023associative,
  title={Associative Memory with Spatio-Temporal Enhancement for Video Anomaly Detection},
  author={Zhong, Yuanhong and Hu, Yongting and Tang, Panliang and Wang, Heng},
  journal={IEEE Signal Processing Letters},
  year={2023},
  publisher={IEEE}
}

@article{li2023multi,
  title={Multi-Branch GAN-based Abnormal Events Detection via Context Learning in Surveillance Videos},
  author={Li, Daoheng and Nie, Xiushan and Gong, Rui and Lin, Ximing and Yu, Hui},
  journal={IEEE Transactions on Circuits and Systems for Video Technology},
  year={2023},
  publisher={IEEE}
}

@article{lu2022learnable,
  title={Learnable locality-sensitive hashing for video anomaly detection},
  author={Lu, Yue and Cao, Congqi and Zhang, Yifan and Zhang, Yanning},
  journal={IEEE Transactions on Circuits and Systems for Video Technology},
  volume={33},
  number={2},
  pages={963--976},
  year={2022},
  publisher={IEEE}
}

@article{fan2025ssim,
	title={SSIM over MSE: A new perspective for video anomaly detection},
	author={Fan, Jin and Chen, Miao and Gu, Zhangyu and Yang, Jiajun and Wu, Huifeng and Wu, Jia},
	journal={Neural Networks},
	volume={185},
	pages={107115},
	year={2025},
	publisher={Elsevier}
}

@inproceedings{gao2025suvad,
	title={SUVAD: Semantic Understanding Based Video Anomaly Detection Using MLLM},
	author={Gao, Shibo and Yang, Peipei and Huang, Linlin},
	booktitle={ICASSP 2025-2025 IEEE International Conference on Acoustics, Speech and Signal Processing (ICASSP)},
	pages={1--5},
	year={2025},
	organization={IEEE}
}

@article{zhong2022bidirectional,
  title={Bidirectional Spatio-Temporal Feature Learning With Multiscale Evaluation for Video Anomaly Detection},
  author={Zhong, Yuanhong and Chen, Xia and Hu, Yongting and Tang, Panliang and Ren, Fan},
  journal={IEEE Transactions on Circuits and Systems for Video Technology},
  volume={32},
  number={12},
  pages={8285--8296},
  year={2022},
  publisher={IEEE}
}

@article{cao2024context,
	title={Context recovery and knowledge retrieval: A novel two-stream framework for video anomaly detection},
	author={Cao, Congqi and Lu, Yue and Zhang, Yanning},
	journal={IEEE Transactions on Image Processing},
	volume={33},
	pages={1810--1825},
	year={2024},
	publisher={IEEE}
}

@inproceedings{liu2023msn,
  title={MSN-net: Multi-Scale Normality Network for Video Anomaly Detection},
  author={Liu, Yang and Li, Di and Zhu, Wei and Yang, Dingkang and Liu, Jing and Song, Liang},
  booktitle={ICASSP 2023-2023 IEEE International Conference on Acoustics, Speech and Signal Processing (ICASSP)},
  pages={1--5},
  year={2023},
  organization={IEEE}
}

@article{liu2023generalized,
	title={Generalized video anomaly event detection: Systematic taxonomy and comparison of deep models},
	author={Liu, Yang and Yang, Dingkang and Wang, Yan and Liu, Jing and Liu, Jun and Boukerche, Azzedine and Sun, Peng and Song, Liang},
	journal={ACM Computing Surveys},
	year={2023},
	publisher={ACM New York, NY}
}

@inproceedings{yang2022dynamic,
	title={Dynamic local aggregation network with adaptive clusterer for anomaly detection},
	author={Yang, Zhiwei and Wu, Peng and Liu, Jing and Liu, Xiaotao},
	booktitle={European Conference on Computer Vision},
	pages={404--421},
	year={2022},
	organization={Springer}
}

@inproceedings{micorek2024mulde,
	title={Mulde: Multiscale log-density estimation via denoising score matching for video anomaly detection},
	author={Micorek, Jakub and Possegger, Horst and Narnhofer, Dominik and Bischof, Horst and Kozinski, Mateusz},
	booktitle={Proceedings of the IEEE/CVF Conference on Computer Vision and Pattern Recognition},
	pages={18868--18877},
	year={2024}
}

@inproceedings{ristea2024self,
	title={Self-distilled masked auto-encoders are efficient video anomaly detectors},
	author={Ristea, Nicolae-C and Croitoru, Florinel-Alin and Ionescu, Radu Tudor and Popescu, Marius and Khan, Fahad Shahbaz and Shah, Mubarak and others},
	booktitle={Proceedings of the IEEE/CVF conference on computer vision and pattern recognition},
	pages={15984--15995},
	year={2024}
}

@inproceedings{ahn2024videopatchcore,
	title={Videopatchcore: An effective method to memorize normality for video anomaly detection},
	author={Ahn, Sunghyun and Jo, Youngwan and Lee, Kijung and Park, Sanghyun},
	booktitle={Proceedings of the Asian Conference on Computer Vision},
	pages={2179--2195},
	year={2024}
}

@article{wang2025icc,
	title={ICC: Intra-Cluster Contraction for Pedestrian Anomaly Detection under One-Class Classification Setting},
	author={Wang, Xiaolei and Wang, Xiaoyang and Bai, Huihui and Lim, Eng Gee and Xiao, Jimin},
	journal={IEEE Transactions on Artificial Intelligence},
	year={2025},
	publisher={IEEE}
}

@article{liu2025crcl,
	title={Crcl: Causal representation consistency learning for anomaly detection in surveillance videos},
	author={Liu, Yang and Wang, Hongjin and Wang, Zepu and Zhu, Xiaoguang and Liu, Jing and Sun, Peng and Tang, Rui and Du, Jianwei and Leung, Victor CM and Song, Liang},
	journal={IEEE Transactions on Image Processing},
	year={2025},
	publisher={IEEE}
}

@article{madan2023self,
	title={Self-supervised masked convolutional transformer block for anomaly detection},
	author={Madan, Neelu and Ristea, Nicolae-C{\u{a}}t{\u{a}}lin and Ionescu, Radu Tudor and Nasrollahi, Kamal and Khan, Fahad Shahbaz and Moeslund, Thomas B and Shah, Mubarak},
	journal={IEEE Transactions on Pattern Analysis and Machine Intelligence},
	volume={46},
	number={1},
	pages={525--542},
	year={2023},
	publisher={IEEE}
}

@inproceedings{zhang2024multi,
	title={Multi-scale video anomaly detection by multi-grained spatio-temporal representation learning},
	author={Zhang, Menghao and Wang, Jingyu and Qi, Qi and Sun, Haifeng and Zhuang, Zirui and Ren, Pengfei and Ma, Ruilong and Liao, Jianxin},
	booktitle={Proceedings of the IEEE/CVF Conference on Computer Vision and Pattern Recognition},
	pages={17385--17394},
	year={2024}
}

@inproceedings{sabokrou2015real,
	title={Real-time anomaly detection and localization in crowded scenes},
	author={Sabokrou, Mohammad and Fathy, Mahmood and Hoseini, Mojtaba and Klette, Reinhard},
	booktitle={Proceedings of the IEEE conference on computer vision and pattern recognition workshops},
	pages={56--62},
	year={2015}
}

@inproceedings{lu2013abnormal,
	title={Abnormal event detection at 150 fps in matlab},
	author={Lu, Cewu and Shi, Jianping and Jia, Jiaya},
	booktitle={Proceedings of the IEEE international conference on computer vision},
	pages={2720--2727},
	year={2013}
}

@article{wang2023video,
	title={Video anomaly detection based on spatio-temporal relationships among objects},
	author={Wang, Yang and Liu, Tianying and Zhou, Jiaogen and Guan, Jihong},
	journal={Neurocomputing},
	volume={532},
	pages={141--151},
	year={2023},
	publisher={Elsevier}
}

@article{li2024stnmamba,
	title={STNMamba: Mamba-based spatial-temporal normality learning for video anomaly detection},
	author={Li, Zhangxun and Zhao, Mengyang and Yang, Xuan and Liu, Yang and Sheng, Jiamu and Zeng, Xinhua and Wang, Tian and Wu, Kewei and Jiang, Yu-Gang},
	journal={arXiv preprint arXiv:2412.20084},
	year={2024}
}

@inproceedings{cheng2023spatial,
	title={Spatial-Temporal Graph Convolutional Network Boosted Flow-Frame Prediction For Video Anomaly Detection},
	author={Cheng, Kai and Zeng, Xinhua and Liu, Yang and Zhao, Mengyang and Pang, Chengxin and Hu, Xing},
	booktitle={ICASSP 2023-2023 IEEE International Conference on Acoustics, Speech and Signal Processing (ICASSP)},
	pages={1--5},
	year={2023},
	organization={IEEE}
}

@inproceedings{yang2023video,
	title={Video Event Restoration Based on Keyframes for Video Anomaly Detection},
	author={Yang, Zhiwei and Liu, Jing and Wu, Zhaoyang and Wu, Peng and Liu, Xiaotao},
	booktitle={Proceedings of the IEEE/CVF Conference on Computer Vision and Pattern Recognition},
	pages={14592--14601},
	year={2023}
}

@article{wu2023dss,
	title={Dss-net: Dynamic self-supervised network for video anomaly detection},
	author={Wu, Peihao and Wang, Wenqian and Chang, Faliang and Liu, Chunsheng and Wang, Bin},
	journal={IEEE Transactions on Multimedia},
	volume={26},
	pages={2124--2136},
	year={2023},
	publisher={IEEE}
}

@article{lyu2025bidirectional,
	title={Bidirectional skip-frame prediction for video anomaly detection with intra-domain disparity-driven attention},
	author={Lyu, Jiahao and Zhao, Minghua and Hu, Jing and Xi, Runtao and Huang, Xuewen and Du, Shuangli and Shi, Cheng and Ma, Tian},
	journal={Pattern Recognition},
	pages={112010},
	year={2025},
	publisher={Elsevier}
}

@article{le2025hstforu,
	title={HSTforU: anomaly detection in aerial and ground-based videos with hierarchical spatio-temporal transformer for U-net},
	author={Le, Viet-Tuan and Jin, Hulin and Kim, Yong-Guk},
	journal={Applied Intelligence},
	volume={55},
	number={4},
	pages={261},
	year={2025},
	publisher={Springer}
}

@inproceedings{wu2020not,
	title={Not Only Look, But Also Listen: Learning Multimodal Violence Detection under Weak Supervision},
	author={Wu, Yuxiang and Zhu, Yao and Guan, Qi and Li, Jun and You, Shaodi and Lu, Tong and Zhang, Weiming},
	booktitle={European Conference on Computer Vision (ECCV)},
	pages={322--339},
	year={2020},
	organization={Springer}
}

@inproceedings{sultani2018real,
	title={Real-world anomaly detection in surveillance videos},
	author={Sultani, Waqas and Chen, Chen and Shah, Mubarak},
	booktitle={Proceedings of the IEEE conference on computer vision and pattern recognition},
	pages={6479--6488},
	year={2018}
}

@inproceedings{yan2023feature,
	title={Feature prediction diffusion model for video anomaly detection},
	author={Yan, Cheng and Zhang, Shiyu and Liu, Yang and Pang, Guansong and Wang, Wenjun},
	booktitle={Proceedings of the IEEE/CVF international conference on computer vision},
	pages={5527--5537},
	year={2023}
}

@article{simonyan2014very,
	title={Very deep convolutional networks for large-scale image recognition},
	author={Simonyan, Karen and Zisserman, Andrew},
	journal={arXiv preprint arXiv:1409.1556},
	year={2014}
}

@article{gu2023mamba,
	title={Mamba: Linear-time sequence modeling with selective state spaces},
	author={Gu, Albert and Dao, Tri},
	journal={arXiv preprint arXiv:2312.00752},
	year={2023}
}

@article{liu2024vmamba,
	title={Vmamba: Visual state space model},
	author={Liu, Yue and Tian, Yunjie and Zhao, Yuzhong and Yu, Hongtian and Xie, Lingxi and Wang, Yaowei and Ye, Qixiang and Jiao, Jianbin and Liu, Yunfan},
	journal={Advances in neural information processing systems},
	volume={37},
	pages={103031--103063},
	year={2024}
}

@inproceedings{hemambaad,
	title={MambaAD: Exploring State Space Models for Multi-class Unsupervised Anomaly Detection},
	author={He, Haoyang and Bai, Yuhu and Zhang, Jiangning and He, Qingdong and Chen, Hongxu and Gan, Zhenye and Wang, Chengjie and Li, Xiangtai and Tian, Guanzhong and Xie, Lei},
	booktitle={The Thirty-eighth Annual Conference on Neural Information Processing Systems},
	year={2024}
}

@inproceedings{huang2022hierarchical,
	title={Hierarchical graph embedded pose regularity learning via spatio-temporal transformer for abnormal behavior detection},
	author={Huang, Chao and Liu, Yabo and Zhang, Zheng and Liu, Chengliang and Wen, Jie and Xu, Yong and Wang, Yaowei},
	booktitle={Proceedings of the 30th ACM international conference on multimedia},
	pages={307--315},
	year={2022}
}

@inproceedings{lyu2025vadmamba,
	author={Lyu, Jiahao and Zhao, Minghua and Hu, Jing and Huang, Xuewen and Chen, Yifei and Du, Shuangli},
	booktitle={2025 IEEE International Conference on Multimedia and Expo (ICME)}, 
	title={VADMamba: Exploring State Space Models for Fast Video Anomaly Detection}, 
	year={2025},
	pages={1-6},
}

@inproceedings{hu2025noise,
	title={Noise-Resistant Video Anomaly Detection via RGB Error-Guided Multiscale Predictive Coding and Dynamic Memory},
	author={Hu, Han and Du, Wenli and Liao, Peng and Wang, Bing and Fan, Siyuan},
	booktitle={Proceedings of the Computer Vision and Pattern Recognition Conference},
	pages={19109--19119},
	year={2025}
}

@article{qiu2024video,
	title={Video anomaly detection guided by clustering learning},
	author={Qiu, Shaoming and Ye, Jingfeng and Zhao, Jiancheng and He, Lei and Liu, Liangyu and Huang, Xinchen and others},
	journal={Pattern Recognition},
	volume={153},
	pages={110550},
	year={2024},
	publisher={Elsevier}
}

@article{vaswani2017attention,
	title={Attention is all you need},
	author={Vaswani, Ashish and Shazeer, Noam and Parmar, Niki and Uszkoreit, Jakob and Jones, Llion and Gomez, Aidan N and Kaiser, {\L}ukasz and Polosukhin, Illia},
	journal={Advances in neural information processing systems},
	volume={30},
	year={2017}
}

@article{park2025fast,
	title={Fast video anomaly detection via context-aware shortcut exploration and abnormal feature distance learning},
	author={Park, Chaewon and Kim, Donghyeong and Cho, MyeongAh and Kim, Minjung and Lee, Minseok and Park, Seungwook and Lee, Sangyoun},
	journal={Pattern Recognition},
	volume={157},
	pages={110877},
	year={2025},
	publisher={Elsevier}
}

@inproceedings{shi2023video,
	title={Video anomaly detection via sequentially learning multiple pretext tasks},
	author={Shi, Chenrui and Sun, Che and Wu, Yuwei and Jia, Yunde},
	booktitle={Proceedings of the IEEE/CVF International Conference on Computer Vision},
	pages={10330--10340},
	year={2023}
}

@inproceedings{luo2017revisit,
	title={A revisit of sparse coding based anomaly detection in stacked rnn framework},
	author={Luo, Weixin and Liu, Wen and Gao, Shenghua},
	booktitle={Proceedings of the IEEE international conference on computer vision},
	pages={341--349},
	year={2017}
}

@article{singh2024attention,
	title={Attention-guided generator with dual discriminator GAN for real-time video anomaly detection},
	author={Singh, Rituraj and Sethi, Anikeit and Saini, Krishanu and Saurav, Sumeet and Tiwari, Aruna and Singh, Sanjay},
	journal={Engineering Applications of Artificial Intelligence},
	volume={131},
	pages={107830},
	year={2024},
	publisher={Elsevier}
}

@article{kommanduri2024dast,
	title={DAST-Net: Dense visual attention augmented spatio-temporal network for unsupervised video anomaly detection},
	author={Kommanduri, Rangachary and Ghorai, Mrinmoy},
	journal={Neurocomputing},
	pages={127444},
	year={2024},
	publisher={Elsevier}
}

@inproceedings{hirschorn2023normalizing,
	title={Normalizing flows for human pose anomaly detection},
	author={Hirschorn, Or and Avidan, Shai},
	booktitle={Proceedings of the IEEE/CVF International Conference on Computer Vision},
	pages={13545--13554},
	year={2023}
}

@article{liang2024c,
	title={C 2 Net: content-dependent and-independent cross-attention network for anomaly detection in videos},
	author={Liang, Jiafei and Xiao, Yang and Zhou, Joey Tianyi and Yang, Feng and Li, Ting and Fang, Zhiwen},
	journal={Applied Intelligence},
	volume={54},
	number={2},
	pages={1980--1996},
	year={2024},
	publisher={Springer}
}

@article{zhong2025two,
  title={A Two-stage Framework with Memory for Anomaly Detection via Video Decomposition and Bidirectional Consistency},
  author={Zhong, Yuanhong and Yan, Ge and Hu, Yongting and Zhu, Dong and Zhu, Ruyue},
  journal={IEEE Transactions on Circuits and Systems for Video Technology},
  year={2025},
  publisher={IEEE}
}

@article{wu2025flow,
	title={DA-flow: Dual attention normalizing flow for skeleton-based video anomaly detection},
	author={Wu, Ruituo and Chen, Yang and Xiao, Jian and Li, Bing and Fan, Jicong and Dufaux, Fr{\'e}d{\'e}ric and Zhu, Ce and Liu, Yipeng},
	journal={IEEE Transactions on Multimedia},
	year={2025},
	publisher={IEEE}
}

@inproceedings{cong2024automatic,
	title={Automatic controllable colorization via imagination},
	author={Cong, Xiaoyan and Wu, Yue and Chen, Qifeng and Lei, Chenyang},
	booktitle={Proceedings of the IEEE/CVF Conference on Computer Vision and Pattern Recognition},
	pages={2609--2619},
	year={2024}
}

%{
%	\small
%	\bibliographystyle{ieeenat_fullname}
%	\bibliography{references}
%}

\end{document}